\documentclass{article} 
\usepackage{iclr2026_conference,times}
\usepackage{afterpage}
\usepackage{hyperref}
\hypersetup{hidelinks}
\usepackage{url}
\usepackage{enumitem}
\usepackage{fvextra}
\usepackage{microtype}
\usepackage{booktabs}
\usepackage{multirow}
\usepackage{tabularx}
\usepackage{array}
\usepackage{graphicx}
\usepackage{amsmath,amssymb}
\usepackage[table]{xcolor}
\usepackage{caption}
\usepackage{pifont}
\usepackage{titletoc}
\usepackage{algorithm}
\usepackage{algpseudocode}
\usepackage{subcaption}
\usepackage{wrapfig}
\usepackage{tcolorbox}
\tcbuselibrary{skins, breakable}
\usepackage{natbib}

\setlist[itemize]{leftmargin=*, nosep}
\setlist[enumerate]{leftmargin=*, nosep}

\newcommand{\cmark}{\ding{51}}

\newcommand{\best}[1]{\textbf{#1}}
\newcommand{\uline}[1]{\underline{#1}}

\definecolor{mygreen}{rgb}{0.27,0.70,0.23}
\definecolor{myyellow}{rgb}{0.74,0.71,0.02}
\definecolor{cadmiumgreen}{rgb}{0.0, 0.42, 0.24}
\definecolor{myred}{rgb}{0.7, 0.3, 0.0}
\definecolor{myblue}{rgb}{0.2, 0.3, 0.6}
\definecolor{promptteal}{HTML}{2B7A78}

\newcolumntype{Y}{>{\raggedright\arraybackslash}X}

\newtcolorbox{promptbox}[2][]{%
  enhanced,
  breakable,
  colback=#2!3,
  colframe=#2!78!black,
  colbacktitle=#2!12,
  coltitle=black,
  title={#1},
  title after break={#1\ (continued)},
  fonttitle=\bfseries,
  boxrule=0.7pt,
  arc=1.5mm,
  left=1.5mm,
  right=1.5mm,
  top=1.2mm,
  bottom=1.2mm,
  before skip=6pt,
  after skip=7pt
}
\newcommand{\promptsection}[1]{%
  \par\smallskip\noindent
  {\bfseries\color{promptteal!75!black}#1}\par\vspace{1pt}
  {\color{promptteal!35}\hrule height 0.4pt}\vspace{2pt}
}
\DefineVerbatimEnvironment{promptverb}{Verbatim}{%
  fontsize=\footnotesize,
  breaklines=true,
  breakanywhere=true
}

\usepackage{amsmath,amsfonts,bm}

\def\eqref#1{equation~\ref{#1}}

\def\1{\bm{1}}

\DeclareMathAlphabet{\mathsfit}{\encodingdefault}{\sfdefault}{m}{sl}
\SetMathAlphabet{\mathsfit}{bold}{\encodingdefault}{\sfdefault}{bx}{n}

\newcommand{\PhaseHeader}[2]{%
  \Statex\vspace{2pt}%
  \colorbox[RGB]{#1}{\parbox{\dimexpr\linewidth-2\fboxsep\relax}{\textbf{#2}}}%
  \vspace{2pt}}
\newcommand{\WhatHeader}[1]{\PhaseHeader{236,232,222}{#1}}
\newcommand{\WhereHeader}[1]{\PhaseHeader{226,232,236}{#1}}
\newcommand{\WhetherHeader}[1]{\PhaseHeader{228,236,228}{#1}}

\vspace{-25pt}
\title{Local Edits, Global Ripples: Replay-Informed Policy Adaptation for Workflow Synthesis}

 \author{%
    \normalfont
    \makebox[\textwidth][c]{Manqing Mao, Hong Wang, Samson Koelle, Jie Yuan, Zhuoer Wang, James Feng,}\\
    \makebox[\textwidth][c]{Yanjun Lin, Daniel Edmiston, Nikki Lijing Kuang, Zhecheng Sheng,
      Wei Niu}\\
    \makebox[\textwidth][c]{Amazon, Inc.}\\
    \makebox[\textwidth][c]{\texttt{\{manqingm, niuwei\}@amazon.com}}
  }

\iclrfinalcopy

\begin{document}

\maketitle

\vspace{-15pt}
\begin{abstract}

Prompt-policy editing offers a practical way to improve agents that synthesize executable workflows without updating the underlying model. However, persistent prompt editing has \textbf{two coupled properties}. First, \emph{edit locality does not imply effect locality}: an edit confined to one predefined policy segment can ripple through downstream execution, altering behavior far beyond the edited segment. Second, \emph{edit effects are composition-sensitive}: edits that work in isolation can interfere after composition, causing one or both to lose their benefit or become harmful. Persistent adaptation must therefore support \textbf{two distinct decisions}: identifying \emph{where} the policy should change from execution feedback, and determining \emph{whether} the resulting edit remains safe to persist after composition. To address these challenges, we introduce \textbf{RIPPLE} (\underline{R}eplay-\underline{I}nformed \underline{P}ersistent \underline{P}olicy \underline{L}ocalization and \underline{E}diting), which addresses these two properties by separating where an edit is made from whether it remains safe after composition. It diagnoses failed trajectories, maps each actionable failure to a predefined policy segment, and restricts the correction to that part of the policy. RIPPLE then evaluates candidates against the same iteration-start policy to compare their isolated gains, before replaying promising edits after previously accepted updates to expose downstream effects and interactions. Only edits that remain safe under composition are retained. We evaluate RIPPLE on \textbf{Flow-HO}, a synthetic held-out benchmark for executable workflow synthesis. RIPPLE improves validation success by up to $23.1\%$ and yields positive gains on two additional frozen language-model backbones, while maintaining strong edit efficiency and low execution cost. Targeted interaction analysis further demonstrates both properties: \textbf{a segment-local tool-use edit changes downstream resource resolution and validation, while an edit beneficial in isolation becomes harmful after composition.}
\end{abstract}

\vspace{-6pt}
\section{Introduction}
\label{sec:intro}

LLM agents are increasingly tasked with producing executable artifacts, such as workflow JSON, infrastructure configurations, and service-integrated automations. Unlike free-form responses, these artifacts must satisfy an execution contract: they must conform to a schema, implement the requested behavior, resolve environment-specific resources, and pass external validation. A workflow can therefore appear semantically plausible but remain undeployable because one field is invalid, one transition is broken, one identifier is stale, or one repair is incomplete.

For frozen-model agents, persistent improvement can target the agent's \emph{prompt policy}: the instructions that govern requirement interpretation, planning, tool use, editing, validation, and repair. Such edits can be reviewed or rolled back without changing weights. However, prompt editing has two coupled properties: \textbf{First, edit locality does not imply effect locality:} a local instruction change can ripple through execution, redirecting tool calls, altering resource resolution and workflow structure, and changing validation or repair behavior~\citep{rippleedits,promptsensitivity}. \textbf{Second, edits can interact under composition:} edits that are beneficial in isolation may interfere once combined, weakening or even reversing one another's gains. Prompt adaptation is therefore a sequential policy-composition problem, not one-shot selection: isolated gains are insufficient evidence for persistence.

Existing methods provide useful mechanisms for proposing and selecting prompt updates. Prompt optimizers search over candidate instructions~\citep{ape,opro,evoprompt,gepa,textgrad}, while recent prompt- and skill-evolution systems use held-out evidence to decide which updates should persist~\citep{skillopt,skillgen,grasp,pace}. However, these mechanisms do not jointly resolve two challenges for executable agents. \textbf{First, execution evidence must be localized into an edit.} A low reward or validation failure identifies an unsuccessful execution, but not the responsible behavior or instruction. Failures in clarification, tool use, schema construction, edit locality, and artifact completeness implicate different prompt segments. \textbf{Second, an edit must be evaluated in the policy state where it will persist.} An edit that improves the base policy can become redundant or harmful after previous updates are composed. \emph{The relevant criterion is therefore not whether an edit works in isolation, but whether it remains useful and safe after composition with the evolving shared policy.}

We introduce \textbf{RIPPLE} (\underline{R}eplay-\underline{I}nformed \underline{P}ersistent \underline{P}olicy \underline{L}ocalization and \underline{E}diting) to address this question. The name reflects both properties above: a local policy edit can \textbf{ripple through the agent's execution flow}, while \textbf{the direction of that ripple can change as the policy evolves}. RIPPLE decomposes adaptation into three decisions. For \emph{what failed}, it assigns trajectory-grounded labels with a deterministic taxonomy. For \emph{where to edit}, it maps each actionable diagnosis to a predefined prompt segment and instantiates a bounded patch from a versioned library, constraining the locus of the policy change. For \emph{whether to persist}, it ranks candidates by local gain and promotes them sequentially, testing whether each local change remains safe after composition. After each acceptance, RIPPLE evaluates the next candidate on top of all previously accepted edits, so downstream interactions are evaluated in the policy state where the candidate would actually persist. The model, tools, and execution environment remain fixed, while accepted patches form auditable, rollbackable checkpoints.

We evaluate RIPPLE on \textbf{Flow-HO}, a synthetic held-out suite for generating executable workflows represented as JSON. RIPPLE consistently improves validation success and composite reward across multiple frozen language models while maintaining strong edit efficiency and low execution cost. Ablations show that replay design and patch composition materially shape the resulting policy, while targeted interaction analysis reveals that local gains can reverse after composition. Together, these results support both aspects of the problem: local edits can have non-local effects, and their value depends on the evolving policy context. In summary, our contributions are:
\begin{enumerate}[label=\textbf{\arabic*.}, leftmargin=*, nosep]

\item \textbf{Failure-localized policy editing.}
We formulate frozen-agent adaptation over a segmented prompt policy, mapping trajectory-grounded diagnoses from a predefined failure taxonomy to target policy segments and bounded corrective edits.

\item \textbf{Composition-aware persistence.}
We introduce a replay-gated promotion procedure that separates local utility from compositional safety by evaluating each candidate against the advancing policy formed by previously accepted edits under separate reward and correctness constraints.

\item \textbf{Flow-HO evaluation and interaction analysis.}
We construct Flow-HO and evaluate RIPPLE across effectiveness, replay design, composition safety, and promotion-rule trade-offs, showing substantial held-out gains and exposing failure modes that arise only through patch interaction.
\end{enumerate}

\section{Persistent Prompt-Policy Adaptation}
\label{sec:formulation}

\subsection{Problem Setup}
\label{sec:problem}

\paragraph{Frozen executable-agent policy.}
We study closed-schema executable workflow synthesis with a frozen,
tool-using LLM agent. Each task $x$ consists of a natural-language
request $q$ and an initial workflow $y_0$, where $y_0=\varnothing$
for generation tasks:
\begin{equation}
\
x=(q,y_0), \qquad
\tau \sim p_\theta(\tau\mid x,\pi), \qquad
\theta \text{ fixed}.
\
\end{equation}
The prompt policy $\pi$ governs the agent's multi-turn interaction
with tools and an external validator. Only $\pi$ is adapted; the model
parameters, tool interface, and execution environment remain fixed.

A completed trajectory $\tau$ receives a bounded composite reward:
\begin{equation}
\
R(\tau;x)
=
\rho\!\left(
S(\tau),C(\tau),E(\tau),K(\tau)
\right),
\
\end{equation}
where $S$ denotes validation success, $C$ is workflow correctness,
$E$ is edit efficiency, and $K$ is normalized execution cost. The exact
task-specific definitions are given in Appendix~\ref{app:reward}.
Because workflow correctness is also constrained separately during
promotion, we retain both $R$ and $C$ as policy-level metrics. For $X\in\{R,C\}$, define the expected metric on task $x$ as
\begin{equation}
\
v_X(\pi;x)
=
\mathbb E_{\tau\sim p_\theta(\cdot\mid x,\pi)}
\left[X(\tau;x)\right].
\
\end{equation}
Given a task pool $\mathcal D$ and an aggregation rule $g$, the pool-level score of policy $\pi$ on metric $X$ is
\begin{equation}
\
V_X^g(\pi;\mathcal D)
=
g\!\left(
\big(v_X(\pi;x)\big)_{x\in\mathcal D}
\right).
\
\end{equation}
Section~\ref{sec:whether} instantiates $g$ as either a uniform mean or a macro-average over workflow families.

\paragraph{Ordered segment-level updates.}
The prompt policy is partitioned into a fixed set of behavioral
segments $\mathcal S$. Let $\mathcal F$ denote a failure taxonomy and
$\mathcal L$ a versioned library of bounded patches, where each
$a\in\mathcal L$ is typed by a failure family $f(a)\in\mathcal F$
and a target segment $s(a)\in\mathcal S$.

An adaptation is an ordered sequence of accepted patches:
\begin{equation}
\
P=(a_1,\ldots,a_t), \qquad
a_j\in\mathcal L, \qquad
\pi_P=\pi_0\oplus P.
\
\end{equation}
The operator $\oplus$ applies patches to their designated segments in
acceptance order. Order matters because an edit may reinforce, override,
or conflict with instructions already present in the policy.

For a candidate patch $a$, an accepted prefix $P$, metric $X$, and task
pool $\mathcal D$, we define the prefix-conditioned marginal effect as:
\begin{equation}
\Delta_X^g(a\mid P;\mathcal D)
=
V_X^g(\pi_0\oplus P\oplus a;\mathcal D)
-
V_X^g(\pi_0\oplus P;\mathcal D).
\label{eq:contextual-delta}
\end{equation}
Thus, the value of a persistent edit depends on both the patch
$a$ and the policy prefix $P$.

\paragraph{Persistent-editing objective.}
Given a training pool $\mathcal D_{\mathrm{train}}$, a replay pool
$\mathcal D_{\mathrm{replay}}$, and a patch budget $H$, the ideal
adaptation objective is:
\begin{equation}
\begin{aligned}
P^\star
\in
\arg\max_{P=(a_1,\ldots,a_t),\,t\le H}
\quad&
V_R(\pi_0\oplus P;\mathcal D_{\mathrm{train}})
\\
\text{s.t.}\quad&
\Delta_R^g
(a_j\mid P^{<j};\mathcal D_{\mathrm{replay}})
\ge \varepsilon_R,
\\
&
\Delta_C^g
(a_j\mid P^{<j};\mathcal D_{\mathrm{replay}})
\ge \varepsilon_C,
\qquad j=1,\ldots,t,
\end{aligned}
\label{eq:persistent-objective}
\end{equation}
where $P^{<j}=(a_1,\ldots,a_{j-1})$. The training objective favors patch sequences that improve performance on the training pool. Before a patch is committed, it must also satisfy reward and correctness thresholds on the replay pool when evaluated on top of all previously accepted patches. Small negative replay deltas is allowed to avoid rejecting potentially useful edits due to rollout variation, but these thresholds are empirical safeguards rather than guarantees on unseen requests.

Solving Eqn.~\ref{eq:persistent-objective} requires to evaluate all ordered subsets of patches up to size $H$. Exhaustive search is prohibitively expensive since each evaluation requires new tool-calling rollouts. Section~\ref{sec:method} therefore approximates this objective with a localized candidate set and a greedy sequential evaluation.

\subsection{Two Decisions in Persistent Policy Editing}
\label{sec:challenges}

Eqn.~\ref{eq:persistent-objective} defines the objective and
constraints for a complete patch sequence, but not how to choose the
next patch. At each step, two decisions remain: which edit should be proposed from a failed execution, and whether that edit should persist after
earlier edits have been accepted.

\vspace{-5pt}

\paragraph{Decision 1: Which edit should be proposed?}
A low reward or validation failure shows that an execution was
unsuccessful, but it does not identify the responsible behavior or the
policy segment that should change. The same scalar outcome may arise
from different failures and therefore require different edits.
Candidate proposal must instead be grounded in execution evidence:
\begin{equation}
\
z(\tau)
\longrightarrow
f
\longrightarrow
s
\longrightarrow
a\in\mathcal L_{f,s},
\
\end{equation}
where $z(\tau)$ denotes trajectory evidence, $f$ a diagnosed failure
family, $s$ the target policy segment, and $a$ a bounded patch.
This decision reduces open-ended prompt rewriting to a localized,
typed update.

\vspace{-5pt}

\paragraph{Decision 2: Should the edit persist?}
An edit that improves the initial policy may become redundant or harmful
once combined with previously accepted edits. The relevant quantity is
therefore its marginal effect on the current policy,
$\Delta_X^g(a\mid P;\mathcal D)$, rather than its isolated effect on
$\pi_0$. Composite reward creates a separate risk: gains in validation,
edit locality, or execution efficiency can offset a decline in workflow
correctness. RIPPLE therefore evaluates reward and correctness separately
before committing an edit.

These two decisions define RIPPLE's structure: proposing an edit is
decomposed into identifying \emph{what} failed and \emph{where} to edit,
while replay determines \emph{whether} the resulting patch should persist. 
\section{RIPPLE: Localized and Replay-Gated Policy Editing}
\label{sec:method}
\begin{figure*}[t]
\centering
\includegraphics[width=1.0\textwidth]{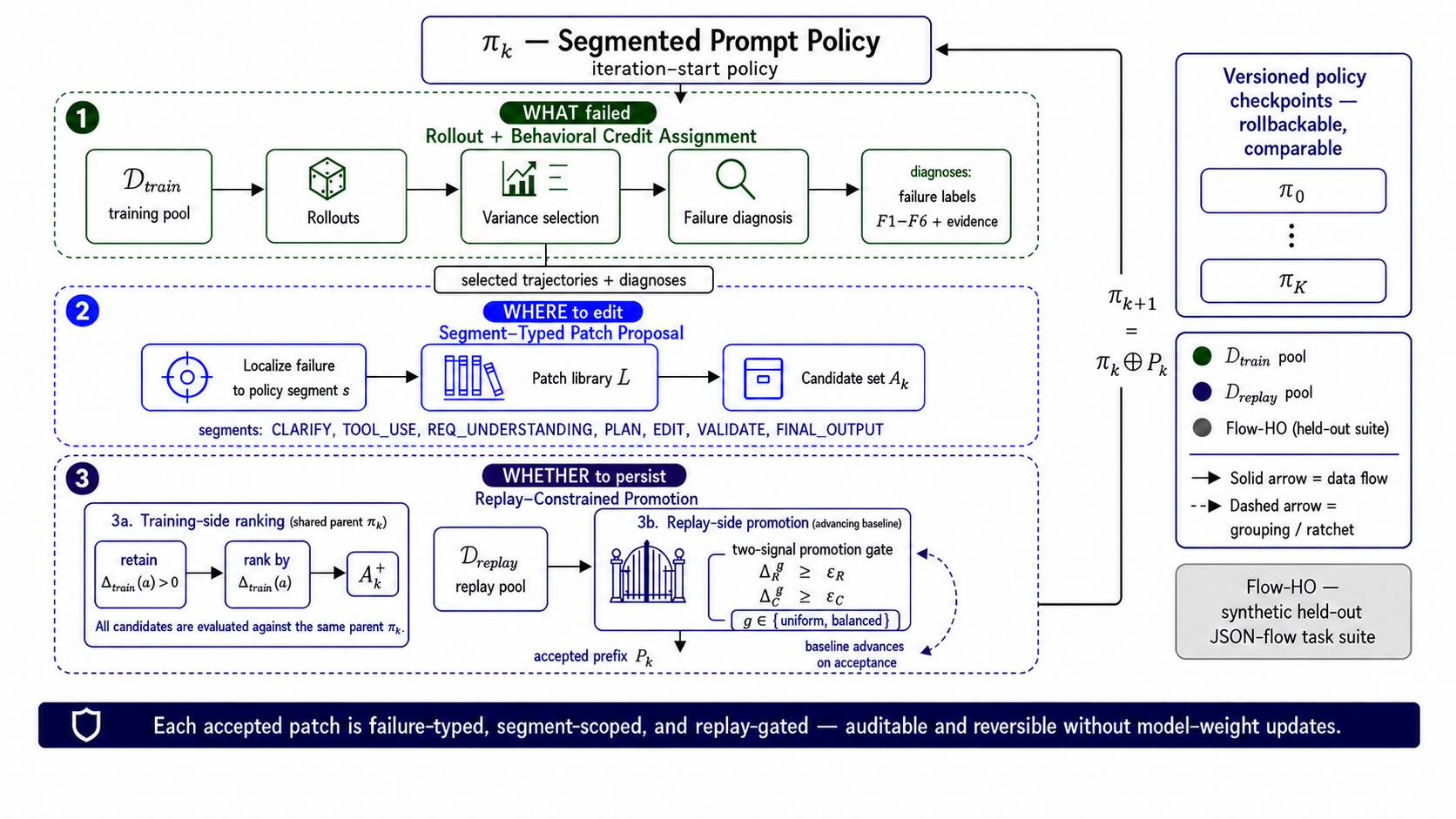}
        \vspace{-38pt}

\caption{\textbf{Overall framework of RIPPLE.}
At iteration $k$, RIPPLE (i) selects informative training rollouts and diagnoses failures with a predefined taxonomy (\textbf{WHAT failed}); (ii) localizes each diagnosis to a policy segment and instantiates bounded patches from a versioned library $\mathcal{L}$ (\textbf{WHERE to edit}); and (iii) ranks candidates by training gain and sequentially evaluates them through a two-signal replay gate against the advancing policy (\textbf{WHETHER to persist}). Accepted patches form the next policy $\pi_{k+1}$. Flow-HO is reserved for final evaluation.}

\label{fig:framework}
\end{figure*}

\vspace{-1pt}
RIPPLE approximates the constrained patch-sequence objective in
Eqn.~\ref{eq:persistent-objective} through greedy sequential policy
improvement over the patch library. Its design separates two scopes of adaptation: localization constrains where a policy change is introduced, while advancing replay evaluates whether the resulting behavioral effects remain safe after composition. At iteration $k$, training
rollouts are used to diagnose failures, construct localized candidate
patches, and rank them by isolated gain. Replay rollouts then evaluate
each candidate on top of the accepted prefix before commitment. The final
evaluation set remains outside the adaptation loop. Figure~\ref{fig:framework} summarizes one adaptation iteration.
Sections~\ref{sec:what}--\ref{sec:where} describe candidate construction,
Section~\ref{sec:whether} describes replay-gated promotion, and
Section~\ref{sec:iteration} describes the outer loop. The full
multi-iteration procedure is given in
Appendix~\ref{app:iteration}.

\subsection{What Failed: Rollout Generation and Behavioral Diagnosis}
\label{sec:what}
\paragraph{Rollout generation.}
For $d\in\mathcal D_{\mathrm{train}}$, RIPPLE samples $N=3$ trajectories under the current policy:
\begin{equation}
\
\tau_{d,n}^{(k)}
\sim
p_\theta(\cdot\mid d,\pi_k),
\qquad n=1,\ldots,N.
\
\end{equation}
Each rollout records the interaction trace, generated workflow, validator feedback, and execution cost.

\paragraph{Verifiable reward.} Each rollout receives a scalar programmatic reward that is a bounded weighted combination of four signals. The weighting and task-specific variants are given in Appendix~\ref{app:reward}:
\begin{equation}
\
R(\tau) = f\big(S(\tau),C(\tau),E(\tau),K(\tau)\big),
\
\end{equation}
where $S \in \{0,1\}$ is service validation success (whether the generated workflow is accepted by the service validator), $C \in [0,1]$ is workflow correctness (whether the workflow implements the intended structure and requested modification), $E \in [0,1]$ is edit efficiency (whether the modification remains within the declared scope), and $K \in [0,1]$ is normalized execution cost. For each datapoint $d$ and policy $\pi$, RIPPLE estimates
$v_X(\pi;d)$ from $N$ independent rollouts:
\begin{equation}
\bar{X}_{\pi}(d)
=
\frac{1}{N}
\sum_{n=1}^{N}
X\left(\tau_{d,n}^{\pi}\right),
\qquad X\in\{R,C\}.
\end{equation}
where $\tau_{d,n}^{\pi}$ is the $n$-th rollout of policy $\pi$ on datapoint $d$. The resulting per-datapoint means are then combined into a pool-level score by the aggregation rule $g$.

\paragraph{Informative-trajectory selection.}
RIPPLE prioritizes for diagnosis the training datapoints with the highest rollout-level reward variance under the current policy. For each $d\in\mathcal D_{\mathrm{train}}$, it computes:
\begin{equation}
\
v_k(d)
=
\operatorname{Var}_{n=1,\ldots,N}
\left[R\!\left(\tau_{d,n}^{(k)}\right)\right].
\
\end{equation}
It then selects the $B_{\mathrm{diag}}$ datapoints with the highest
reward variance and diagnoses the lowest-reward rollout from each selected datapoint. High variance indicates inconsistent behavior under the current policy, while the lowest-reward rollout provides a concrete failure trace for
diagnosis.
\begin{equation}
\
\tau_d^-
=
\arg\min_{n\in\{1,\ldots,N\}}
R\!\left(\tau_{d,n}^{(k)}\right).
\
\end{equation}

\vspace{-10pt}
\paragraph{Predicate-based behavioral diagnosis.}
For each selected trajectory $\tau$, RIPPLE applies predefined rules
to the execution log, checking for signals such as missing clarification,
failed tool use, validator errors, ineffective repair, out-of-scope
changes, and incomplete output. The rules return the actionable failure
families and their supporting evidence, without an LLM call:
\begin{equation}
\
\operatorname{Diagnose}(\tau)
=
\bigl(\mathcal{F}(\tau), z(\tau)\bigr),
\qquad
\mathcal{F}(\tau)\subseteq\mathcal{F},
\
\end{equation}
where $z(\tau)$ records the triggered predicates and supporting trace
evidence, and $\mathcal{F}(\tau)$ contains the actionable failure
families supported by that evidence. Multiple families may be activated
because an upstream error can produce several downstream symptoms, or
because distinct failures co-occur in the same trajectory. For
attribution, fixed precedence rules designate one activated family as
primary, preferring direct and causally earlier evidence; the remaining
actionable labels are retained and may seed additional candidate
patches. Full predicates and precedence rules are given in
Appendix~\ref{app:taxonomy}.

RIPPLE uses six failure families: \textbf{F1 Clarification} (missing required information), \textbf{F2 Tool use} (missing, incorrect, or misordered tool calls), \textbf{F3 Schema} (schema or service-constraint violations), \textbf{F4 Repair behavior} (failure to act on validator feedback), \textbf{F5 Edit locality} (out-of-scope modifications), and \textbf{F6 Completeness} (partial or unresolved artifacts). The primary label captures the earliest causal failure, while secondary labels retain additional actionable symptoms.  When multiple labels are plausible, RIPPLE prefers the earliest causal failure in the execution trace over the final visible symptom. Full disambiguation rules and the evidence schema are provided in Appendix~\ref{app:taxonomy}.

\subsection{Where to Edit: Segment-Typed Patch Construction}
\label{sec:where}

\paragraph{Policy segmentation.}
Before adaptation begins, the base prompt is partitioned into seven
behavioral segments:
\texttt{REQ\_UNDERSTANDING}, \texttt{CLARIFY}, \texttt{PLAN},
\texttt{TOOL\_USE}, \texttt{EDIT}, \texttt{VALIDATE}, and
\texttt{FINAL\_OUTPUT}.
Each segment governs a distinct stage of agent execution. Holding boundaries fixed keeps localization consistent across iterations and gives every patch a stable and reviewable target. RIPPLE adapts the policy only by appending bounded instructions to designated segments. 

\paragraph{Failure-localized patch construction.}
For each selected trajectory $\tau$ and each diagnosed failure family
$f\in\mathcal F(\tau)$, RIPPLE deterministically selects a target segment
$s$ and retrieves a patch $a$:
\begin{equation}
\
s=\operatorname{Localize}\!\left(f,z(\tau)\right),
\qquad
a=\operatorname{Retrieve}\!\left(\mathcal L,f,s\right).
\
\end{equation}
The patch library $\mathcal L$ contains predefined entries indexed by failure
family and target segment. Given a diagnosed failure $f$ localized to segment
$s$, RIPPLE first retrieves the entry matching the $(f,s)$ pair. If no exact
match exists, it falls back to a generic entry targeting the same segment.
The retrieved instruction is appended verbatim to $s$; no LLM is invoked.
Each library entry stores a patch ID, failure family, target segment, and
instruction, making the procedure deterministic and reproducible.

\paragraph{Candidate aggregation and coverage ordering.}
Let $\mathcal P(\tau)$ denote the set of patches nominated by a selected
trajectory $\tau$. The support count of patch $a$ at iteration $k$ is
\begin{equation}
\
c_k(a)
=
\sum_{\tau\in\mathcal T_k}
\mathbf 1\!\left[a\in\mathcal P(\tau)\right],
\
\end{equation}
where $\mathcal T_k$ denotes the trajectories selected for diagnosis.
The count $c_k(a)$ measures how many independent failure traces nominate
patch $a$, so a larger count indicates stronger support for that update. RIPPLE merges nominations by patch ID, removes previously accepted patches,
and ranks candidates within each failure family by $c_k(a)$. It then applies
a coverage-first round robin, allowing each represented family to contribute
one candidate before any family contributes a second. The resulting ordered
candidate set proceeds to training-side evaluation.

\subsection{Whether to Persist: Policy Improvement and Replay-Constrained Promotion}
\label{sec:whether}

RIPPLE separates candidate ranking from commitment. Training-side
pre-qualification evaluates every patch from the same iteration-start
policy $\pi_k$, making their one-step gains directly comparable.
Replay-side promotion then evaluates candidates sequentially on top of
previously accepted edits, testing each patch in the policy state where
it would actually persist. This separation preserves fair ranking while testing whether a local edit's downstream effects remain safe after composition.
\vspace{-5pt}
\paragraph{Training-side policy improvement.}
Each candidate patch $a\in\mathcal A_k$ defines a candidate policy
$\pi_k^a=\pi_k\oplus a$. To compare candidates on a common basis,
RIPPLE evaluates every patch against the same iteration-start policy
$\pi_k$, isolating its one-step effect from interactions with other
candidates. The resulting one-step training gain is:
\begin{equation}
\
\Delta_{\mathrm{train}}(a)
=
\widehat V_R(\pi_k^a;\mathcal D_{\mathrm{train}})
-
\widehat V_R(\pi_k;\mathcal D_{\mathrm{train}}),
\
\end{equation}
where
$\widehat V_R(\pi;\mathcal D)
=
|\mathcal D|^{-1}\sum_{d\in\mathcal D}\bar R_\pi(d)$
is the empirical mean reward over the task pool.
Only positive-gain candidates advance:
$\mathcal A_k^+
=
\{a\in\mathcal A_k:\Delta_{\mathrm{train}}(a)>0\}$,
and RIPPLE sorts them by decreasing $\Delta_{\mathrm{train}}(a)$.
This stage identifies promising edits and determines their evaluation
order; replay-side promotion decides whether they remain safe after
composition.
\vspace{-5pt}
\paragraph{Replay-side policy promotion.}
After sorting $\mathcal A_k^+$ by decreasing
$\Delta_{\mathrm{train}}$, let $a_j$ denote the $j$-th candidate patch
in this order. Let $P_k^{<j}$ denote the ordered sequence of candidates
accepted among the first $j-1$ positions. The policy used to evaluate
$a_j$ and its candidate extension are
$\pi_k^{<j}=\pi_k\oplus P_k^{<j}$ and
$\tilde{\pi}_{k,j}=\pi_k^{<j}\oplus a_j$.
RIPPLE first averages the $N=3$ rollouts for each replay datapoint. It then computes a single replay score from these per-datapoint means using one of two weighting schemes:

\begin{itemize}[leftmargin=*, nosep]
    \item \textbf{Uniform aggregation} $g_{\mathrm{uniform}}$:
    all replay datapoints receive equal weight;

    \item \textbf{GT-family-balanced aggregation} $g_{\mathrm{balanced}}$
    (\emph{coverage-weighted} in labels):
    datapoints are averaged within each ground-truth workflow
    family, then equally across represented families
    (Appendix~\ref{app:coverage-terminology}).
\end{itemize}

Candidate $a_j$ is evaluated marginally against the latest accepted
policy:
\begin{equation}
\
\Delta_R^g(a_j\mid P_k^{<j})
=
g\!\left[\bar R_{\tilde{\pi}_{k,j}}\right]
-
g\!\left[\bar R_{\pi_k^{<j}}\right],
\qquad
\Delta_C^g(a_j\mid P_k^{<j})
=
g\!\left[\bar C_{\tilde{\pi}_{k,j}}\right]
-
g\!\left[\bar C_{\pi_k^{<j}}\right].
\
\end{equation}
Candidate $a_j$ is accepted only when both replay deltas satisfy their
tolerances:
\begin{equation}
\
\operatorname{Accept}(a_j\mid P_k^{<j})
\iff
\Delta_R^g(a_j\mid P_k^{<j})\geq\varepsilon_R
\ \land\
\Delta_C^g(a_j\mid P_k^{<j})\geq\varepsilon_C.
\
\end{equation}
We use $\varepsilon_R=-0.05$ and $\varepsilon_C=-0.10$. If $a_j$ is
accepted, $\tilde{\pi}_{k,j}$ becomes the policy used to evaluate the
next candidate; if rejected, the current policy remains unchanged.
Intuitively, each accepted patch changes the policy context in which
later patches are judged. Replay therefore evaluates each candidate against the current accepted policy, making downstream effects and interactions with earlier edits observable before persistence. The gate constrains each patch's replay
regression rather than cumulative change from $\pi_k$; because the
tolerances are negative, small regressions are allowed and may accumulate
across patches. Checking correctness separately guards against patches
that pass the reward threshold despite a loss in workflow
correctness. Appendix~\ref{app:correctness-binding} analyzes when this
constraint binds, while Appendix~\ref{app:gate-design} discusses the
AND rule and tolerance choices.

\subsection{Policy Iteration and Replay Maintenance}
\label{sec:iteration}

The patches accepted at iteration $k$ form the ordered sequence
$P_k=(a_{k,1},\ldots,a_{k,m_k})$, and the next iteration starts from
$\pi_{k+1}=\pi_k\oplus P_k$. If no patch is accepted,
$P_k=\varnothing$ and $\pi_{k+1}=\pi_k$. Thus, each outer iteration
carries forward only edits that pass replay-side promotion.

RIPPLE supports two replay modes: (i) \textbf{Static replay} always uses
$\mathcal D_{\mathrm{replay}}=\mathcal D_{\mathrm{core}}$; and (ii)
\textbf{Adaptive replay} uses
$\mathcal D_{\mathrm{replay}}^{(k)}
=\mathcal D_{\mathrm{core}}
\cup\mathcal D_{\mathrm{hard}}^{(k)}
\cup\mathcal D_{\mathrm{recent}}^{(k)}$.
After scoring the iteration-$k$ training rollouts and before promotion,
RIPPLE refreshes the $\mathcal D_{\mathrm{hard}}^{(k)}$ with persistently low-performing
examples and the $\mathcal D_{\mathrm{recent}}^{(k)}$ with examples whose success status
changed. The fixed core provides a stable reference, the refreshed
subsets track current failure modes; the resulting replay
set gates that iteration's candidates.

An iteration with no accepted patch increments a patience counter; any
acceptance resets it. RIPPLE stops after two consecutive no-update
iterations. Full algorithm is shown in Appendix~\ref{app:iteration}.


\begin{figure}[H]
\centering
\includegraphics[width=1.02\linewidth]{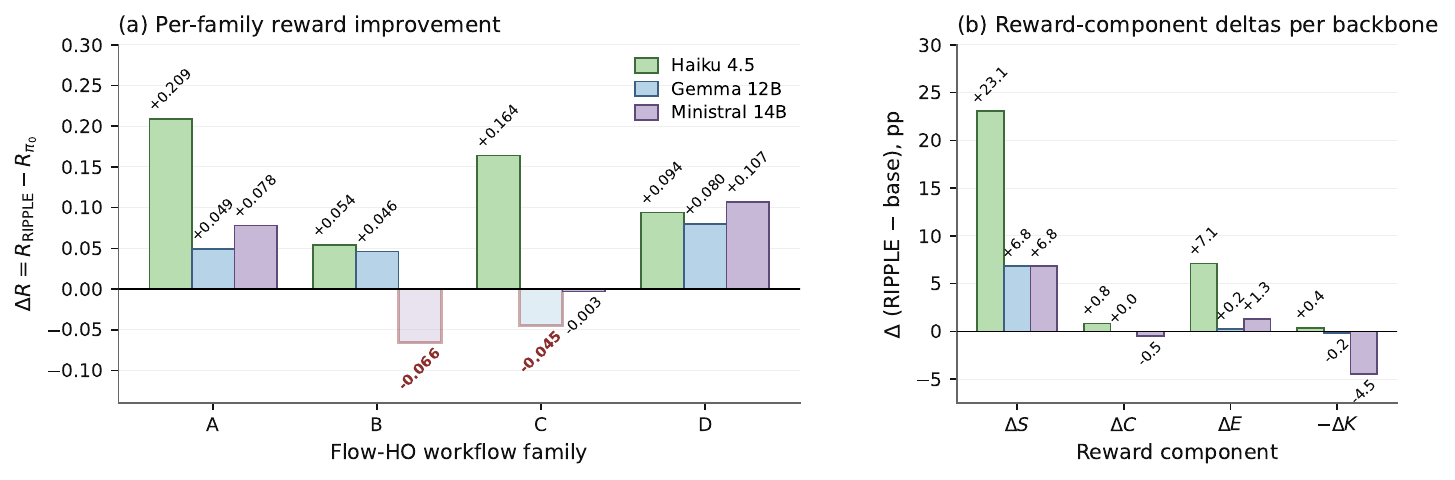}
\vspace{-15pt}
\caption{Cross-backbone reward improvement on Flow-HO modification. \textbf{(a)} Per-family paired $\Delta R = R_{\mathrm{RIPPLE}} - R_{\pi_0}$: every backbone improves on at least three of four families, and the two regressions (red-outlined) are backbone-specific---no family regresses across all three models. \textbf{(b)} Reward-component deltas (percentage points; $K$ negated so up is better): the gain is carried by service-validation success ($\Delta S \geq +6.8$~pp on every backbone; $+23.1$~pp on Haiku), while workflow correctness ($\Delta C$) is flat across backbones ($|\Delta C| \leq 0.9$~pp). Haiku uses RIPPLE-U; Gemma and Ministral use transfer runs with $N=3$ per task.}

\label{fig:cross-family-delta}
\end{figure}

\section{Experiments}
\label{sec:experiments}

In this section, we examine whether RIPPLE improves held-out workflow quality
across frozen policy backbones and whether replay supports safe edit persistence
under iterative composition. We first describe the experimental setup and then
organize the evaluation for 4 research questions (RQs):

\noindent\textbf{RQ1. Cross-backbone effectiveness.}
Does RIPPLE improve held-out workflow quality across frozen policy backbones
with different capability levels?

\noindent\textbf{RQ2. Replay design.}
How do aggregation, replay maintenance, gating, and iteration depth jointly
affect patch acceptance, policy evolution, and the quality of the selected policy?

\noindent\textbf{RQ3. Composition safety.}
Can the advancing replay baseline detect and reject edits that are beneficial
in isolation but harmful after composition with previously accepted edits?

\noindent\textbf{RQ4. Promotion rules.}
How do stricter acceptors affect policy composition and quality?

\subsection{Experimental Setup}
\label{sec:setup}

\paragraph{Flow-HO.}
Flow-HO is a synthetic held-out benchmark of executable workflow tasks in a JSON-based flow language. Its primary split contains 39 workflow-modification tasks, where the agent must modify a corrupted workflow according to a request and pass external validation. The four ground-truth families are disjoint from training and replay and are never used for adaptation or model selection. Appendix~\ref{app:benchmarks} provides details and the secondary workflow-generation setting.

\paragraph{Systems.}
Claude Haiku 4.5 is the primary frozen policy backbone for RQ1–RQ4; Gemma 3-12B-it and Ministral 3-14B provide two lower-capability transfer settings (Appendix B.5.1). Transfer backbones evaluate only base $\pi_0$ and RIPPLE. No variant selection is performed on transfer backbones. On Haiku, we compare $\pi_0$ with five RIPPLE variants: \textbf{RIPPLE-U} (uniform, adaptive replay), \textbf{RIPPLE-S} (family-balanced, static replay), \textbf{Coverage-Adaptive} (family-balanced, adaptive replay), \textbf{Train-Only} (no replay promotion), and \textbf{Multi-Iteration}. RQ1 refers to RIPPLE-U simply as \textbf{RIPPLE}. RQ4 additionally compares controlled SkillOpt-style and SkillGen-style acceptors with candidate trajectories held fixed. Appendix~\ref{app:baselines} provides adaptation and baseline details.

\paragraph{Protocol.}
We use $N=3$ rollouts per datapoint and set the diagnosis budget to
$B_{\mathrm{diag}}=15$. We report service-validation success, workflow correctness, and composite reward. Haiku deltas are paired against $\pi_0$ over the same 39 tasks, with $10{,}000$ family-cluster bootstrap replicates; Gemma and Ministral use paired point estimates. Edit efficiency, execution cost, model settings are reported in Appendices~\ref{app:reward-breakdown} and \ref{app:model-config}.

\subsection{RQ1: Held-Out Effectiveness across Frozen Backbones}
\label{sec:principal}
\label{sec:consistency}

\vspace{-8pt}

\noindent
\begin{minipage}[t]{0.52\linewidth}
\vspace{0pt}
Table~\ref{tab:main} reports the primary cross-backbone comparison; the full Haiku design study appears in \S\ref{sec:ablation}:

\vspace{5pt}

\textbf{RIPPLE primarily improves deployability on the principal backbone.}
On Haiku~4.5, RIPPLE raises service-validation success from $54.7\%$ to $77.8\%$ ($+23.1\%$) and composite reward from $0.525$ to $0.644$; both paired family-cluster intervals exclude zero. Correctness rises by only $0.8\%$ and remains unresolved. The main effect is a higher rate of workflows that survive service validation rather than a broad shift in structural similarity.

\medskip
\textbf{The gain remains efficient and spans the benchmark families.}
RIPPLE also attains the strongest edit efficiency and lowest execution cost among the harmonized systems. Its mean reward exceeds the base policy in all four Flow-HO families. Appendix~\ref{app:reward-breakdown} provides the family-level results.
\end{minipage}
\hfill
\begin{minipage}[t]{0.45\linewidth}
\vspace{0pt}
\centering
\setlength{\tabcolsep}{3.7pt}
\renewcommand{\arraystretch}{0.98}
\footnotesize
\begin{tabular}{@{}lccc@{}}
\toprule
\textbf{Policy}
  & \textbf{Success (\%)}
  & \textbf{Corr. (\%)}
  & \textbf{Reward} \\
\midrule
\rowcolor[RGB]{240,240,240}
\multicolumn{4}{c}{\textit{Claude Haiku 4.5 (primary backbone)}} \\
Base $\pi_0$
  & 54.7 & 74.9 & 0.525 \\
RIPPLE
  & \best{77.8}$^{\dagger}$ & \best{75.7} & \best{0.644}$^{\dagger}$ \\
\addlinespace[1pt]
\rowcolor[RGB]{240,240,240}
\multicolumn{4}{c}{\textit{Gemma 3-12B-it (transfer backbone)}} \\
Base $\pi_0$
  & 37.6 & \best{80.7} & 0.460 \\
RIPPLE
  & \best{44.4} & \best{80.7} & \best{0.492} \\
\addlinespace[1pt]
\rowcolor[RGB]{240,240,240}
\multicolumn{4}{c}{\textit{Ministral 3-14B (transfer backbone)}$^{\ast}$} \\
Base $\pi_0$
  & 24.8 & \best{59.7} & 0.299 \\
RIPPLE
  & \best{31.6} & 59.2 & \best{0.327} \\
\bottomrule
\end{tabular}
\captionsetup{hypcap=false}
\vspace{-5pt}
\captionof{table}{Flow-HO modification results across different frozen backbones. RIPPLE improves validation success and reward in all three settings. $\dagger$ marks a paired family-cluster 95\% interval excluding zero (Appendix B.3.5); $^*$ marks the Ministral protocol (Appendix B.5.5). Best means are bold.}
\label{tab:main}

\end{minipage}
\par\medskip
\vspace{-10pt}

\paragraph{Positive untuned gains transfer to both weaker backbones.} RIPPLE raises service-validation success by $6.8$~pp on both Gemma and Ministral and improves reward on both, while correctness is unchanged on Gemma and slightly lower on Ministral. Accepted patches are largely backbone-specific, consistent with diagnosis adapting to distinct failure profiles (Appendix~\ref{app:cross-model-patches}). Crucially, both runs reuse Haiku's proposer, patch library, replay design, and gate tolerances without backbone-specific tuning. \textbf{These positive gains therefore demonstrate untuned transfer, and the residual Haiku--transfer gap should not be interpreted as a method ceiling.} Ministral only bypasses the compute-saving positive-gain pre-filter; replay gating is unchanged (Appendix~\ref{app:ministral-protocol}). We treat these point estimates as transfer evidence rather than a precise cross-backbone ranking.

\vspace{-4pt}

\paragraph{Gains are broad across families and driven by service validation.}
Figure~\ref{fig:cross-family-delta}(a) shows positive reward gains on all four families for Haiku and three of four for each transfer backbone, with regressions occurring on different families. Panel~(b) shows that service-validation success drives most of the gain ($\Delta S=+23.1,+6.8,+6.8$~pp), while workflow correctness remains essentially flat ($|\Delta C|\leq0.9$~pp). Full results are in Appendices~\ref{app:family-reward-cross} and~\ref{app:reward-breakdown-cross}.

\vspace{-6pt}

\begin{table}[H]
\centering
\setlength{\tabcolsep}{7pt}
\renewcommand{\arraystretch}{0.96}
\footnotesize
\begin{tabular}{@{}l c c c@{}}
\toprule
\textbf{Method}
  & \textbf{Success (\%)}
  & \textbf{Corr. (\%)}
  & \textbf{Reward} \\
\midrule
RIPPLE-S {\small(static replay)}
  & \uline{76.1}\,{\scriptsize\textcolor{myred}{$\downarrow \phantom{1}1.7$}}
  & \best{77.9}\,{\scriptsize\textcolor{cadmiumgreen}{$\uparrow 2.2$}}
  & \uline{0.639}\,{\scriptsize\textcolor{myred}{$\downarrow 0.005$}} \\
Coverage-Adaptive {\small($g_{\mathrm{bal}}$ + adaptive)}
  & 69.2\,{\scriptsize\textcolor{myred}{$\downarrow \phantom{1}8.5$}}
  & \uline{76.8}\,{\scriptsize\textcolor{cadmiumgreen}{$\uparrow 1.1$}}
  & 0.599\,{\scriptsize\textcolor{myred}{$\downarrow 0.045$}} \\
Train-Only {\small(no replay gate)}
  & 71.8\,{\scriptsize\textcolor{myred}{$\downarrow \phantom{1}6.0$}}
  & 75.6\,{\scriptsize\textcolor{myred}{$\downarrow 0.1$}}
  & 0.613\,{\scriptsize\textcolor{myred}{$\downarrow 0.031$}} \\
Multi-Iteration {\small($k>1$)}
  & 64.1\,{\scriptsize\textcolor{myred}{$\downarrow 13.7$}}
  & 74.3\,{\scriptsize\textcolor{myred}{$\downarrow 1.4$}}
  & 0.571\,{\scriptsize\textcolor{myred}{$\downarrow 0.074$}} \\
\midrule
\rowcolor[RGB]{225,240,225}
RIPPLE-U {\small(uniform + adaptive)}
  & \best{77.8}
  & 75.7
  & \best{0.644} \\
\bottomrule
\end{tabular}
\vspace{-6pt}
\caption{Haiku design study on Flow-HO modification. Arrows show changes from RIPPLE-U; RQ2 contrasts each ablation with Coverage-Adaptive. \best{Bold} and \uline{underline} mark the top two raw means.}
\label{tab:ablation}
\end{table}
\vspace{-10pt}

\subsection{RQ2: Replay Design and Iteration Depth}
\label{sec:ablation}

\afterpage{%
\noindent
\begin{minipage}[t]{0.455\linewidth}
\vspace{0pt}
\centering
\setlength{\tabcolsep}{3.5pt}
\renewcommand{\arraystretch}{1.02}
\footnotesize
\begin{tabular}{@{}l l@{}}
\toprule
\textbf{Context} & \textbf{Observed effect} \\
\midrule
Shared parent $\pi_0$
  & $\Delta_{\mathrm{train}}=+0.0465$ \\
\midrule
\multirow{3}{*}{Accepted prefix $P$}
  & $\Delta_R^g=-0.2257$ \\
  & $\Delta_C^g=-0.1218$ \\
  & Success: $0.354\rightarrow 0.000$ \\
\bottomrule
\end{tabular}
\captionsetup{hypcap=false}
\captionof{table}{F2a flips after composition: positive on $\pi_0$, destructive after prefix $P$.}
\label{tab:interaction}

\end{minipage}
\hfill
\begin{minipage}[t]{0.5\linewidth}
\vspace{0pt}
   \centering
    \setlength{\tabcolsep}{3.2pt}
    \renewcommand{\arraystretch}{1.05}
    \footnotesize
    \begin{tabular}{@{}l c l@{}}
    \toprule
    \textbf{Acceptor} & \textbf{Patches} & \textbf{$S/C/R$} \\
    \midrule
    RIPPLE-U (headline)
      & 4 & \textbf{77.8}/75.7/\textbf{0.644} \\
    \midrule
    RIPPLE $\varepsilon$-AND
      & 4 & 69.2/76.8/0.599 \\
    Skill-Compact
      & 1 & 70.9/\textbf{82.2}/0.625 \\
    \bottomrule
    \end{tabular}
    \captionsetup{hypcap=false}
    \vspace{-8pt}
    \captionof{table}{Controlled acceptor swap. Rows~2--3 share one trajectory, differing only in acceptor. Skill-Compact is the one-patch checkpoint $\pi_0\oplus\text{F3a}$ for both the SkillOpt and SkillGen.}
    \label{tab:acceptor}

\end{minipage}
\par\medskip
\vspace{3pt}
}

\paragraph{Replay design governs which policy persists.}
Table~\ref{tab:ablation} ablates aggregation, replay maintenance, gating, and iteration depth around Coverage-Adaptive. Uniform aggregation and static replay both outperform Coverage-Adaptive in success and reward, with the matched static-versus-adaptive success interval excluding zero, showing that aggregation and replay maintenance change checkpoint selection rather than merely its evaluation. Train-Only reaches similar aggregate performance but accepts a broader patch set, including edits rejected by replay; thus, similar held-out means can mask materially different persistent policies, and replay gating controls which local gains are allowed to compose into the shared policy. More adaptation is likewise not inherently better: Multi-Iteration adds two patches yet underperforms its one-iteration parent, while checkpoints with the same patch count can differ substantially in reward. \textbf{These results show that persistent performance is determined by which edits survive, their order, and their interactions, not simply by patch count or iteration depth.} Appendices~\ref{app:convergence-diagnostics} and ~\ref{app:patch-profile} provide the patch decisions and convergence traces.

\vspace{-4pt}
\subsection{RQ3: Detecting Destructive Edit Interactions}
\label{sec:interaction-rq}

\paragraph{Composition can reverse a locally beneficial edit.}
To answer RQ3, we isolate F2a, which restricts resource lookup to newly introduced references (Table~\ref{tab:interaction}). Against the shared iteration-start policy $\pi_0$, F2a passes training-side pre-qualification with $\Delta_{\mathrm{train}}=+0.0465$. After four earlier patches are composed, however, the same edit becomes strongly destructive: replay reward and correctness fall by $0.2257$ and $0.1218$, respectively, while service-validation success drops from $35.4\%$ to $0\%$. The reversal is interaction-induced: an earlier patch requires concrete resource resolution, whereas F2a exempts identifiers already present in the corrupted input, allowing stale values to survive into publication. Advancing-baseline replay exposes this failure and rejects the patch before persistence. \textbf{Thus, edit quality is policy-context dependent: an edit that is beneficial against $\pi_0$ can become harmful after composition.} Appendix~\ref{app:interaction-case} provides the full trace-level analysis.

\vspace{-4pt}
\subsection{RQ4: Promotion Rules and Checkpoint Trade-offs}
\label{sec:acceptor}

\paragraph{Promotion rules select distinct policy-composition trade-offs.}
To answer RQ4, we hold the proposer, candidate order, and replay evidence fixed and vary only the acceptor (Table~\ref{tab:acceptor}). The controlled swap uses the Coverage-Adaptive trajectory, while RIPPLE-U remains the strongest checkpoint and the headline reference in Table~\ref{tab:main}. RIPPLE's tolerant two-signal gate accepts four patches, whereas the SkillOpt-style~\citep{skillopt} and SkillGen-style~\citep{skillgen} rules both retain only F3a, converging on the same one-patch checkpoint that we call \textbf{Skill-Compact} ($\pi_0\oplus\text{F3a}$); this isolates promotion stringency from proposal quality. Skill-Compact improves correctness by $5.4$~pp and reward by $0.026$ over the four-patch $\varepsilon$-AND checkpoint (paired $95\%$ intervals $[+1.7,+9.4]$ and $[+0.003,+0.051]$), with only a $1.7\%$ difference in success. Yet these gains are specific to the matched acceptor probe: RIPPLE-U remains the best overall configuration in the main evaluation, while both checkpoints win on subsets of requests, indicating that acceptance stringency changes not only checkpoint size but also where gains accrue. A post hoc oracle further reveals selector headroom. \textbf{Thus, promotion is a substantive policy-construction decision: stricter acceptors can produce more compact checkpoints and improve certain metrics, but they reshape rather than uniformly improve the resulting policy and do not supersede RIPPLE-U as the strongest overall configuration.} Appendix~\ref{app:checkpoint-complementarity} provides request-level and selector analyses.

\vspace{-6pt}

\section{Conclusion}
\label{sec:conclusion}
\vspace{-2pt}
We introduced \textbf{RIPPLE}, a replay-informed framework for persistent prompt-policy adaptation in executable workflow synthesis. RIPPLE turns execution failures into bounded, segment-localized edits and promotes each candidate only after replay against the evolving accepted policy. On Flow-HO, RIPPLE substantially improves service-validation success on Claude Haiku 4.5 and transfers consistently across two additional frozen backbones, while preserving correctness and achieving strong edit efficiency at low execution cost. Mechanistic analyses reveal that local gains can reverse after composition and that promotion rules select distinct checkpoints even under a fixed proposer and replay evidence. Together, these findings establish persistent prompt adaptation as a sequential policy-composition problem: the value of an edit depends not only on its local utility, but on how it interacts with the policy in which it persists. \textbf{In this sense, a local policy edit can create global ripples through downstream execution, making compositional safety central to persistent adaptation.}

\bibliographystyle{iclr2026_conference}
\bibliography{references}

\clearpage
\appendix
\raggedbottom

\section*{Appendix Contents}
\startcontents[appendix]
\begingroup
\setlength{\parskip}{4pt}
\titlecontents{section}
  [0em]
  {\vspace{4pt}\bfseries}
  {\contentslabel{2.0em}}
  {}
  {\titlerule*[0.5pc]{.}\contentspage}
\titlecontents{subsection}
  [2.0em]
  {\vspace{2pt}}
  {\contentslabel{3.2em}}
  {}
  {\titlerule*[0.5pc]{.}\contentspage}
\titlecontents{subsubsection}
  [5.2em]
  {\vspace{1pt}\small}
  {\contentslabel{4.2em}}
  {}
  {\titlerule*[0.5pc]{.}\contentspage}
\printcontents[appendix]{}{1}{}
\endgroup

\section{Related Work}
\label{app:related-work}

RIPPLE lies at the intersection of automatic prompt optimization, verification-driven improvement, and persistent skill or prompt evolution. Related methods share a generate--evaluate--promote pattern; RIPPLE differs in how it localizes execution evidence, represents edits, and evaluates them after composition with an evolving policy.

We organize this discussion around the two properties from Section~\ref{sec:intro}. First, \textbf{edit locality does not imply effect locality}: a local textual change can ripple through downstream behavior, as documented in the knowledge-editing~\citep{rippleedits} and prompt-sensitivity~\citep{promptsensitivity} literatures. Second, \textbf{the value of an edit is policy-context dependent}: whether an improvement survives depends on the accepted prefix already in force. The subsections below thread these two properties through prompt optimization (A.1), verification-driven improvement (A.2), persistent skill evolution (A.3), and executable workflow synthesis (A.4), highlighting where each family addresses one property, both, or neither.

\subsection{Automatic Prompt and Policy Optimization}

\paragraph{Search and textual optimization.} APE~\citep{ape} and OPRO~\citep{opro} optimize candidate instructions from task scores, while ProTeGi/APO~\citep{protegi} and TextGrad~\citep{textgrad} turn natural-language feedback into targeted textual updates. EvoPrompt~\citep{evoprompt} and GEPA~\citep{gepa} use evolutionary refinement, whereas DSPy/MIPRO~\citep{dspy,mipro} and SAMMO~\citep{sammo} optimize instructions, demonstrations, or structured metaprompt fragments within LLM programs. Collectively, these methods show that black-box textual policies can improve without gradient access to the underlying model. Because they score each candidate against a task metric on the base policy alone, however, they do not test whether an edit remains safe after other edits have been composed. The observation that small textual changes can produce non-local behavior shifts~\citep{rippleedits,promptsensitivity} is precisely what motivates evaluating a candidate in the advancing policy state, not only against $\pi_0$.

\paragraph{Search, reflection, and evolutionary proposers.} A parallel line refines prompts through explicit search or self-referential rewriting: PromptAgent~\citep{promptagent} runs Monte Carlo tree search over error-reflection actions to reach expert-level prompts, Promptbreeder~\citep{promptbreeder} co-evolves task and mutation prompts under LLM-graded fitness, and SPO~\citep{spo} removes the need for reference labels by comparing candidate outputs with a pairwise LLM judge. Agent Symbolic Learning~\citep{agentsymbolic} extends the textual-gradient view to whole-agent parameters (prompts, tools, and pipeline structure), and Trace / OptoPrime~\citep{trace} generalizes the pattern by treating execution traces and rich textual feedback as ``gradients'' that an LLM optimizer applies to heterogeneous parameters, including prompts and code. Both optimize the current parameter jointly against a task loss; neither retains a versioned accepted prefix against which a new candidate must remain safe, so gains that depend on the current parameter state are not distinguished from gains that would compose across accepted edits. RIPPLE shares the search-and-reflect intuition but replaces LLM-driven proposal with deterministic retrieval from a fixed patch library indexed by predicate diagnosis; unbounded proposal, in our setting, expands the search space faster than the replay gate can safely filter it.

\paragraph{Relation to RIPPLE.} RIPPLE follows the same broad propose-and-select template but constrains proposal before evaluation: diagnosed execution evidence selects a predefined policy segment and a bounded library patch. It also decouples ranking from persistence. Shared-parent training gains rank candidates, whereas replay tests each candidate after previously accepted edits have been composed. A patch can therefore be useful in isolation yet unsafe in the policy state in which it would persist.

\paragraph{Model-weight adaptation.} Weight-updating methods such as GRPO~\citep{grpo} optimize model parameters from rollout outcomes. RIPPLE instead keeps the model endpoint fixed and adapts only the prompt policy, so accepted changes remain inspectable, versionable, and reversible.

\subsection{Verification-Driven and Conservative Improvement}

External verification is often necessary for reliable improvement. Intrinsic self-correction can reduce accuracy without oracle feedback~\citep{huang-selfcorrect}, whereas verify-then-revise systems such as CRITIC~\citep{critic} use tools or external checks to repair individual outputs. RIPPLE extends the same verify-before-commit principle from a single response to a persistent policy edit.

The replay gate also parallels conservative policy improvement in offline RL~\citep{cql}, but applies an empirical constraint: RIPPLE withholds a training-improving candidate when replay reward or workflow correctness exceeds the allowed regression under the advancing accepted prefix. Automated correctness signals make repeated promotion practical, and prior LLM-as-a-judge work suggests that model-based evaluation can correlate with human preferences~\citep{llmjudge}. RIPPLE nevertheless combines automated correctness with programmatic service validation rather than relying on an LLM judge alone.

\paragraph{Verifiable rewards and iterative acceptance.} A related family scales verification-driven improvement to full model training. Reinforcement Learning with Verifiable Rewards~\citep{tulu3} and DeepSeek-R1~\citep{deepseekr1} update model weights against deterministic checkers (math answers, unit tests) rather than preference models; rStar-Math~\citep{rstarmath} advances a base policy in rounds gated by a learned process-preference model over trajectories; and learned verifiers such as V-STaR~\citep{vstar} and Generative Verifiers~\citep{genrm} train models to rerank generations at inference. RIPPLE inherits the same verifiable-signal premise but stays at the symbolic-policy layer: no weight updates, no learned verifier, no best-of-$N$ reranking of outputs. Verification enters only at promotion time, as a two-signal replay constraint that a candidate patch must pass against an advancing accepted prefix before it enters the deployed policy.

\subsection{Persistent Skill and Prompt Evolution}

The idea of persisting textual artifacts across tasks has an older lineage. Voyager~\citep{voyager} maintains a lifelong library of executable skills paired with an automatic curriculum and self-verification loop; Reflexion~\citep{reflexion} treats per-episode natural-language reflections as verbal reinforcement over a fixed policy; ExpeL~\citep{expel} accumulates free-form insights from training tasks and retrieves them at inference; ADAS~\citep{adas} lets an LLM meta-agent program new agents into an ever-growing archive; Agent Workflow Memory~\citep{awm} induces reusable workflows from browser trajectories and calls them back on later web tasks; and AutoGuide~\citep{autoguide} distills state-conditioned natural-language guidelines from offline experience and retrieves the applicable ones at test time. Both AWM's workflow-level artifacts and ADAS's agent-graph edits produce persistent behavior changes whose downstream effects are non-local, and whose value can shift once earlier artifacts are already in place. AutoGuide is the closest external analog to RIPPLE's persistent guideline layer, but it selects guidelines by state at inference rather than gating their entry into the deployed policy against an advancing accepted prefix. Together, these systems establish that acquired text can carry non-trivial behavior across tasks, but none couple acquisition to a replay-based falsifiability rule against the current accepted prefix---any generated skill, reflection, guideline, or workflow that improves the demonstration is retained.

The closest methods to RIPPLE also use explicit promotion rules for persistent textual updates:
\begin{itemize}[leftmargin=*, itemsep=1pt, topsep=2pt]
    \item \textbf{GRASP}~\citep{grasp} induces an open-vocabulary failure taxonomy, proposes ADD/MODIFY/REMOVE skill edits, and promotes candidates when fixes exceed regressions under a hard regression budget.
    \item \textbf{SkillGen}~\citep{skillgen} compares paired rollouts with and without a skill and promotes it when repair gains exceed a minimum paired-count threshold.
    \item \textbf{SkillOpt}~\citep{skillopt} optimizes a monolithic skill file and accepts a candidate only when it strictly improves a disjoint selection score; ties are rejected.
    \item \textbf{PACE}~\citep{pace} uses paired outcomes and an anytime-valid e-process for sequential acceptance on a reused development pool, together with a fresh audit pool.
\end{itemize}

Replay-gated promotion is therefore not unique to RIPPLE. The distinction lies in its deployment target and the structure surrounding the gate. First, RIPPLE targets schema-validated executable workflows, for which service validity, structural correctness, edit locality, resource resolution, and execution cost can diverge. Second, diagnosis uses a fixed six-family workflow taxonomy rather than a run-specific open vocabulary. Third, every update is typed by both failure family and prompt segment and retrieved from a versioned library. Finally, promotion constrains reward and workflow correctness separately while evaluating candidates against an advancing accepted prefix. These choices trade open-ended coverage for deterministic, auditable policy changes suited to persistent deployment.

\subsection{Executable Workflow Synthesis and Verifiable-Reward Agents}

RIPPLE evaluates on schema-validated executable workflows, and its verifier stack (service validity, structural correctness, resource resolution, execution) mirrors a growing class of code-agent benchmarks. SWE-bench~\citep{swebench} grades LLM patches against hidden unit tests on real GitHub issues, and $\tau$-Bench~\citep{taubench} exercises multi-turn tool-using dialogue against a simulated user and environment with pass$^k$ execution scoring; both establish the ``deterministic checker as reward'' regime that Flow-HO adapts to workflow synthesis. On the agent side, SWE-agent~\citep{sweagent} wraps an LLM in an edit--run--observe loop over a live repository, CodeAct~\citep{codeact} unifies actions as executable code so that interpreter feedback drives per-step revision, and Self-Debug~\citep{selfdebug} shows that executable feedback alone---compiler and interpreter output, with no human critique---suffices for intra-episode repair. Closer to persistent workflow construction, AFlow~\citep{aflow} searches over code-represented workflow graphs with Monte Carlo tree search to synthesize whole pipelines under a task metric, but the search yields a single deployed workflow rather than a versioned sequence of typed prompt-segment edits. In all of these systems, executable feedback is consumed only within the current task or a one-shot search: the prompt policy is fixed at deployment, so any signal extracted from one instance is discarded when the next arrives. RIPPLE occupies a complementary slot in this landscape---the same class of executable signals is retained across tasks by admitting only those prompt-segment edits that pass the two-signal replay gate against an advancing baseline, converting intra-episode feedback into cross-task, versioned policy state.

\section{Implementation Details and Additional Experiments}
\label{app:impl}

\subsection{Benchmarks}
\label{app:benchmarks}

\paragraph{Flow-HO.} Flow-HO is a synthetic held-out suite of executable workflow tasks expressed in a JSON-based flow language. It covers workflow generation and modification. Every example was created for this study from publicly documented flow constructs; no customer workflow or interaction data is included.

The primary study evaluates \textbf{workflow modification}: the agent receives a natural-language request and a corrupted workflow $y_0$, then must return a validated workflow that implements the requested change. \textbf{Workflow generation} is secondary: the agent receives only the request and constructs a schema-valid workflow from scratch.

\subsubsection{Workflow Modification}

The modification benchmark contains 12 ground-truth workflow families partitioned into disjoint training, replay-core, and Flow-HO pools (Table~\ref{tab:pools}). A family contains all corruption variants derived from one ground-truth workflow, and no family crosses pool boundaries.

\begin{table}[t]
\centering
\setlength{\tabcolsep}{6pt}
\renewcommand{\arraystretch}{1.12}
\footnotesize
\begin{tabular}{l r r}
\toprule
\textbf{Pool} & \textbf{Datapoints} & \textbf{Ground-truth families} \\
\midrule
$\mathcal{D}_{\mathrm{train}}$ & 35 & 5 \\
$\mathcal{D}_{\mathrm{core}}$ & 8 & 3 \\
Flow-HO (modification) & 39 & 4 \\
\bottomrule
\end{tabular}
\vspace{-5pt}
\caption{Modification-pool composition. Ground-truth workflow families are disjoint across training, replay core, and Flow-HO.}
\label{tab:pools}
\end{table}

\paragraph{Held-out families.} The four Flow-HO modification families are denoted A--D. Table~\ref{tab:families} summarizes their task counts and topology. \textbf{Blocks} counts workflow actions, \textbf{types} counts distinct action types, and \textbf{transitions} counts base, conditional, and error \texttt{NextAction} edges. The families span distinct structural regimes rather than minor variants of one template, supporting the family-level analysis in \S\ref{sec:consistency}.

\begin{table*}[t]
\centering
\setlength{\tabcolsep}{4pt}
\renewcommand{\arraystretch}{1.08}
\scriptsize
\begin{tabularx}{\textwidth}{@{}c c c p{0.28\textwidth} X@{}}
\toprule
\textbf{Family} & $n$ & \textbf{B/T/E} & \textbf{Dominant pattern} & \textbf{Structural character} \\
\midrule
A & 9 & 9/8/13 & Mixed; no action type appears more than twice
  & Short linear voice flow with logging, greeting, input capture, function lookup, one comparison branch, and disconnect. \\
B & 10 & 10/6/16 & Three voice updates and three message actions
  & Three-language switch whose localized greeting branches reconverge before a routing transfer. \\
C & 10 & 17/10/39 & Five comparisons and four routing-target updates
  & Largest, most branch-dense flow: channel gate, function check, comparison cascade, wait loop, and four-way routing. \\
D & 10 & 14/10/30 & Five session-attribute updates
  & Attribute-heavy onboarding with recording, knowledge-session initialization, bot handoff, and routing transfer. \\
\bottomrule
\end{tabularx}
\vspace{-6pt}
\caption{Structural properties of the four Flow-HO modification families. B/T/E denotes the numbers of blocks, distinct action types, and transitions; transitions include base, conditional, and error edges.}
\label{tab:families}
\end{table*}

\paragraph{Corruption procedure.} Each corrupted input is generated deterministically while preserving a resolvable graph and a valid start action. The operator names denote the requested repair: \texttt{add-block} removes one action and redirects incoming edges; \texttt{reroute} changes one outgoing or error transition; \texttt{modify-config} overwrites one parameter or metadata field; and \texttt{replace-logic} removes a self-contained subgraph and reconnects the surviving path. Metadata for deleted actions are removed, and every remaining edge resolves to an existing action.

Across the 82 modification datapoints (35 training, 8 replay core, and 39 Flow-HO), the corruption mix is 25 \texttt{add-block}, 19 \texttt{modify-config}, 19 \texttt{reroute}, and 19 \texttt{replace-logic} examples. Pool-specific counts are stored with the benchmark metadata.

\subsubsection{Workflow Generation}

The secondary generation benchmark contains 49 synthetic workflows with 8--25 actions. Each datapoint stores a natural-language request, a ground-truth workflow, a coarse complexity label, a scenario category, and metadata for missing-information slots and refusal conditions. Generation uses the same agent and validation infrastructure, but correctness is computed with the identifier-invariant metric in Appendix~\ref{app:reward}. Edit efficiency is fixed at one because no starting workflow is modified.

\subsection{Baselines and Controlled Acceptors}
\label{app:baselines}

\textbf{Base $\pi_0$} is a human-authored segmented prompt policy created once with assistance from a stronger reference model and then frozen. Every RIPPLE configuration starts from the same $\pi_0$ and changes it only through accepted library patches.

RQ4 changes only the acceptor, holding RIPPLE's proposer, candidate order, and recorded replay outcomes fixed. The \textbf{SkillOpt-style} rule~\citep{skillopt} accepts only when replay reward strictly improves, $\Delta_R^g>0$; ties are rejected. The \textbf{SkillGen-style} rule~\citep{skillgen} accepts when
\[
\operatorname{round}(mN\,\Delta_S^g)
\geq
\max\{2,\lceil0.05mN\rceil\},
\]
where $m=|\mathcal D_{\mathrm{replay}}|$ and $N$ is the rollout count per replay datapoint. These controlled acceptor swaps are not end-to-end reproductions of the original systems. Both select the same one-patch checkpoint, $\pi_0\oplus\mathrm{F3a}$, which we call \textbf{Skill-Compact}. Its metrics come from a separate matched acceptor-probe cache and are analyzed in RQ4 rather than ranked with the harmonized systems in Table~\ref{tab:main}.

\subsection{Framework Details}
\label{app:framework}

This subsection specifies the deterministic components summarized in Section~\ref{sec:method}: diagnosis and candidate construction (Appendix~\ref{app:candidate-construction}), replay-gated promotion and replay-pool maintenance (Appendix~\ref{app:promotion}), the multi-iteration procedure (Appendix~\ref{app:iteration}), the reward function (Appendix~\ref{app:reward}), and rollout and uncertainty settings (Appendix~\ref{app:model-config}).

\begin{table}[H]
\centering
\footnotesize
\setlength{\tabcolsep}{3.2pt}
\renewcommand{\arraystretch}{1.18}
\begin{tabularx}{\linewidth}{@{}>{\bfseries}p{0.07\linewidth}p{0.13\linewidth}Yp{0.14\linewidth}@{}}
\toprule
Family & Failure & Predicate evidence & Default target \\
\midrule
F1 & Clarification & \textbf{Direct:} \path{missed_clarification_slots}; \path{clarification_loop}. & \texttt{CLARIFY} \\
F2 & Tool use & \textbf{Direct:} \path{tool_errors}; \path{no_tool_calls}. \textbf{Fallback:} \path{no_resource_resolution_tools}. & \texttt{TOOL\_USE} \\
F3 & Schema & \textbf{Direct:} \path{hallucinated_action_types}; \path{validation_errors}. \textbf{Fallback:} \path{low_correctness} when $S=0$ and $C<0.3$. & \texttt{EDIT} \\
F4 & Repair & \textbf{Direct:} \path{repair_attempted_but_failed}, defined as more than one validation call while $S=0$. & \texttt{VALIDATE} \\
F5 & Edit locality & \textbf{Direct:} \path{low_efficiency}. \textbf{Fallback:} \path{partial_correctness_but_failed}. & \texttt{EDIT} \\
F6 & Completeness & \textbf{Direct:} \path{partial_output}; \path{missing_required_features} for generation. \textbf{Fallback:} \path{no_generated_flow}. & \texttt{EDIT} (subfamily overrides) \\
\bottomrule
\end{tabularx}
\vspace{-6pt}
\caption{Predicate-based failure taxonomy for behavioral credit assignment. Direct evidence precedes consequence fallbacks; F6 refinements can override the default target segment.}
\label{tab:failure-taxonomy}
\end{table}

\subsubsection{Candidate Construction: Diagnosis, Library, and Aggregation}
\label{app:candidate-construction}
\label{app:taxonomy}
\label{app:library}

\paragraph{Taxonomy and library provenance.} We constructed the taxonomy \emph{top-down} from the major actionable failure loci in executable workflow synthesis, rather than inducing it from Flow-HO outcomes. Intuitively, a failed trajectory can arise because the agent did not obtain the needed information (F1), interact with tools or resources correctly (F2), produce a schema-valid artifact (F3), recover from validation feedback (F4), preserve the intended scope of an edit (F5), or complete the requested workflow (F6). These categories correspond to distinct intervention points in the agent policy, making the taxonomy useful for localizing \emph{where} corrective credit should be assigned rather than merely describing observed errors.

The six super-families F1--F6 therefore predate the Flow-HO evaluation used in this paper and were fixed before the final held-out split was constructed. Their tie-break orderings reflect task-aware causal priority rather than empirical prevalence in training rollouts; for example, clarification and tool-use failures take precedence when editing an existing flow. The F6 subrefinements F6a--F6h were introduced iteratively during method development: F6a--F6d were derived from an initial reward-component profile of low-reward $\pi_0$ training rollouts; F6e and F6f targeted missing-block and edge failures observed in subsequent adaptation runs; F6g and F6h were added after an iteration in which no candidate was accepted, targeting the parameter-key and edge failure modes identified in that trace. Every library entry was frozen before its own replay-gate evaluation, and no accepted patch was rewritten after its replay outcome. The 23 entries were authored by the paper authors with reference to the public flow-schema documentation, and retrieval invokes no LLM.

\paragraph{Predicate-based diagnosis.} For each selected low-reward trajectory $\tau$, RIPPLE returns an actionable failure-family set $\mathcal F(\tau)$ and an evidence record $z(\tau)$. Direct predicates identify observed causal behavior; consequence predicates capture downstream symptoms and are considered only when direct evidence is unavailable. Table~\ref{tab:failure-taxonomy} summarizes the six families and their default target segments.

When multiple predicates fire, direct evidence takes precedence. Remaining ties follow a task-specific order: F1, F2, F3, F4, F5, F6 for modification, and F3, F6, F2, F1, F4, F5 for generation. A hallucinated action type overrides either order with F3, F6, F1, F2, F4, F5. The first family is marked primary; other supported families remain actionable and may nominate additional patches.

\paragraph{Completeness refinement.} F6 is refined using the reward-component trace: no output maps to F6a/\texttt{PLAN}; missing blocks to F6e/\texttt{PLAN}; missing requested capabilities to F6b/\texttt{REQ\_UNDERSTANDING}; unresolved values to F6c/\texttt{TOOL\_USE}; incorrect edges to F6f/\texttt{EDIT}; and residual under-specification to F6d/\texttt{EDIT}. F6g is a second preregistered requirement-coverage patch targeting \texttt{REQ\_UNDERSTANDING}. Every refinement retrieves a fixed library entry; none triggers free-form patch generation.

\paragraph{Segment-typed patch library.} The base prompt is parsed into seven fixed segments---\texttt{REQ\_UNDERSTANDING}, \texttt{CLARIFY}, \texttt{PLAN}, \texttt{TOOL\_USE}, \texttt{EDIT}, \texttt{VALIDATE}, and \texttt{FINAL\_OUTPUT}---covering request interpretation, clarification, planning, resource resolution, workflow editing, validation and repair, and response completion. The default family-to-segment map is $F1\!\to\!\texttt{CLARIFY}$, $F2\!\to\!\texttt{TOOL\_USE}$, $F3\!\to\!\texttt{EDIT}$, $F4\!\to\!\texttt{VALIDATE}$, $F5\!\to\!\texttt{EDIT}$, and $F6\!\to\!\texttt{EDIT}$; the F6 refinements above override this default.

The versioned library $\mathcal L$ contains 23 preregistered entries of the form $(\text{id},f,s,\text{instruction})$. Given family $f$ and localized segment $s$, \textsc{Retrieve} first selects an entry matching both and otherwise falls back to an entry for $s$. The stored instruction is appended verbatim to the target segment. Retrieval invokes no LLM, so each accepted edit is deterministic, reviewable, and independently reversible.

\paragraph{Aggregation.} Repeated nominations are merged by patch ID, and previously accepted IDs are removed. The support count $c_k(a)$ records how many selected trajectories nominate patch $a$. Candidates are sorted by support within each failure family, then interleaved by a coverage-first round robin: every represented family contributes one candidate before any family contributes a second. The leading candidates form $\mathcal A_k$ and proceed to promotion.

\subsubsection{Promotion: Two-Signal Replay Gate and Replay Pool}
\label{app:promotion}
\label{app:gate-detail}
\label{app:replay-pool}

\paragraph{Empirical scores and replay deltas.} For policy $\pi$, metric $X\in\{R,C\}$, and pool $\mathcal D$, define
\[
\widehat V_X^g(\pi;\mathcal D)
=
 g\!\left((\bar X_{\pi}(d))_{d\in\mathcal D}\right),
\qquad
\bar X_{\pi}(d)=\frac{1}{N}\sum_{n=1}^{N}X(\tau_{d,n}^{\pi}).
\]
Uniform aggregation weights datapoints equally. Ground-truth-family-balanced aggregation first averages within each workflow family and then weights represented families equally. The implementation label \emph{coverage-weighted} refers to this macro-average, not to coverage over diagnosed failure families (Appendix~\ref{app:coverage-terminology}).

For the $j$-th ranked candidate $a_j$, the empirical replay delta corresponding to Equation~\ref{eq:contextual-delta} is
\[
\widehat\Delta_X^g(a_j\mid P_k^{<j};\mathcal D_{\mathrm{replay}}^{(k)})
=
\widehat V_X^g(\pi_k\oplus P_k^{<j}\oplus a_j;\mathcal D_{\mathrm{replay}}^{(k)})
-
\widehat V_X^g(\pi_k\oplus P_k^{<j};\mathcal D_{\mathrm{replay}}^{(k)}).
\]
A candidate is promoted only when both deltas satisfy $(\varepsilon_R,\varepsilon_C)=(-0.05,-0.10)$. Acceptance advances the prefix; rejection leaves it unchanged. Train-Only skips replay promotion after training-side pre-qualification.

\paragraph{Replay-pool composition.} The promotion pool is assembled once per outer iteration, after scoring the parent-policy training rollouts and before evaluating candidates. Adaptive replay uses
\[
\mathcal D_{\mathrm{replay}}^{(k)}
=
\mathcal D_{\mathrm{core}}
\cup\mathcal D_{\mathrm{hard}}^{(k)}
\cup\mathcal D_{\mathrm{recent}}^{(k)},
\]
with duplicates removed. The fixed core contains eight datapoints spanning three ground-truth workflow families. The hard slot contains up to four non-core training datapoints with parent-policy mean reward below $0.4$ and at least two failed rollouts, ordered from lowest reward upward. The recent slot contains up to four non-core datapoints whose task-success status changed between $\pi_{k-1}$ and $\pi_k$; any unfilled positions are backfilled from the low-reward pool. At iteration~1, the recent slot is initialized by the same backfill rule.

Static replay uses only $\mathcal D_{\mathrm{core}}$ and omits both adaptive slots. RIPPLE-U and Coverage-Adaptive use adaptive replay but different aggregators; RIPPLE-S combines static replay with balanced aggregation. Because adaptive replay may include training datapoints, it is a mixed stability pool rather than a fully disjoint validation set. Flow-HO remains outside the adaptation loop.

\subsubsection{Iteration Procedure}
\label{app:iteration}

Algorithm~\ref{alg:ripple-single-iter} specifies one outer iteration, including replay construction, shared-parent candidate ranking, and advancing-prefix promotion. Algorithm~\ref{alg:ripple-full} composes accepted sequences across iterations, checkpoints after every iteration, and stops after two consecutive iterations with no accepted patch.

\begin{algorithm}[p]
\caption{One RIPPLE adaptation iteration}
\label{alg:ripple-single-iter}
\small
\setlength{\baselineskip}{10pt}
\begin{algorithmic}[1]
\setlength{\itemsep}{1.2pt}
\Require Current policy $\pi_k$; training pool $\mathcal D_{\mathrm{train}}$;
         replay core $\mathcal D_{\mathrm{core}}$; replay mode
         $m\in\{\textsc{Static},\textsc{Adaptive}\}$ and history $H_{k-1}$;
         patch library $\mathcal L$; aggregator $g$; rollout count $N$;
         diagnosis budget $B_{\mathrm{diag}}$; tolerances $(\varepsilon_R,\varepsilon_C)$;
         previously accepted patch IDs $A_{<k}$.
\Ensure Next policy $\pi_{k+1}$; accepted sequence $P_k$; updated history $H_k$;
        iteration record $\mathcal R_k$.

\WhatHeader{Phase 1 --- WHAT failed: rollout generation and diagnosis}
\ForAll{$d\in\mathcal D_{\mathrm{train}}$}
    \State Sample $\tau_{d,n}^{(k)}\sim p_\theta(\cdot\mid d,\pi_k)$ for $n=1,\ldots,N$
    \State $\bar X_{\pi_k}(d)\gets N^{-1}\sum_{n=1}^{N}X(\tau_{d,n}^{(k)})$
           for $X\in\{R,C\}$
    \State $v_k(d)\gets\operatorname{Var}_{n}[R(\tau_{d,n}^{(k)})]$;
           $\tau_d^-\gets\arg\min_n R(\tau_{d,n}^{(k)})$
\EndFor
\State $H_k\gets\textsc{UpdateReplayHistory}(H_{k-1},\{\tau_{d,n}^{(k)}\}_{d,n})$
\If{$m=\textsc{Static}$}
    \State $\mathcal D_{\mathrm{replay}}^{(k)}\gets\mathcal D_{\mathrm{core}}$
\Else
    \State $(\mathcal D_{\mathrm{hard}}^{(k)},\mathcal D_{\mathrm{recent}}^{(k)})
           \gets\textsc{RefreshReplay}(H_k)$
    \State $\mathcal D_{\mathrm{replay}}^{(k)}\gets
           \mathcal D_{\mathrm{core}}\cup\mathcal D_{\mathrm{hard}}^{(k)}
           \cup\mathcal D_{\mathrm{recent}}^{(k)}$
\EndIf
\State $\mathcal T_k\gets\{\tau_d^-:d\in\operatorname{Top}_{B_{\mathrm{diag}}}(v_k)\}$
\ForAll{$\tau\in\mathcal T_k$}
    \State $(\mathcal F(\tau),z(\tau))\gets\textsc{Diagnose}(\tau)$
\EndFor

\WhereHeader{Phase 2 --- WHERE to edit: segment-typed patch construction}
\ForAll{$\tau\in\mathcal T_k$}
    \State $\mathcal P(\tau)\gets\varnothing$
    \ForAll{$f\in\mathcal F(\tau)$}
        \State $s\gets\textsc{Localize}(f,z(\tau))$;
               $a\gets\textsc{Retrieve}(\mathcal L,f,s)$
        \State $\mathcal P(\tau)\gets\mathcal P(\tau)\cup\{a\}$
    \EndFor
\EndFor
\State $c_k(a)\gets\sum_{\tau\in\mathcal T_k}\mathbf 1[a\in\mathcal P(\tau)]$
       for each nominated patch $a$
\State $\mathcal A_k\gets\textsc{CoverageFirstOrder}
       (\{a\in\bigcup_{\tau}\mathcal P(\tau):\operatorname{id}(a)\notin A_{<k}\},c_k)$

\WhetherHeader{Phase 3 --- WHETHER to persist: shared-parent ranking and replay promotion}
\ForAll{$a\in\mathcal A_k$}
    \State $\pi_k^a\gets\pi_k\oplus a$
    \State $\Delta_{\mathrm{train}}(a)\gets
      \widehat V_R(\pi_k^a;\mathcal D_{\mathrm{train}})-
      \widehat V_R(\pi_k;\mathcal D_{\mathrm{train}})$
\EndFor
\State $\mathcal A_k^+\gets\{a\in\mathcal A_k:\Delta_{\mathrm{train}}(a)>0\}$,
       ordered by decreasing $\Delta_{\mathrm{train}}$
\State $P_k\gets()$; $\mathcal R_k\gets\varnothing$
\For{$j=1,\ldots,|\mathcal A_k^+|$}
    \State $a_j\gets\mathcal A_k^+[j]$; $P_k^{<j}\gets P_k$
    \State $\pi_k^{<j}\gets\pi_k\oplus P_k^{<j}$;
           $\widetilde\pi_{k,j}\gets\pi_k^{<j}\oplus a_j$
    \For{$X\in\{R,C\}$}
        \State $\Delta_X^g(a_j\mid P_k^{<j};\mathcal D_{\mathrm{replay}}^{(k)})\gets
        \widehat V_X^g(\widetilde\pi_{k,j};\mathcal D_{\mathrm{replay}}^{(k)})-
        \widehat V_X^g(\pi_k^{<j};\mathcal D_{\mathrm{replay}}^{(k)})$
    \EndFor
    \If{$\Delta_R^g(a_j\mid P_k^{<j})\geq\varepsilon_R$
        \textbf{and} $\Delta_C^g(a_j\mid P_k^{<j})\geq\varepsilon_C$}
        \State $P_k\gets P_k\mathbin\Vert(a_j)$ \Comment{advance the replay baseline}
        \State Log $(a_j,\Delta_R^g,\Delta_C^g,\textsc{Accept})$ in $\mathcal R_k$
    \Else
        \State Log $(a_j,\Delta_R^g,\Delta_C^g,\textsc{Reject})$ in $\mathcal R_k$
    \EndIf
\EndFor
\State $\pi_{k+1}\gets\pi_k\oplus P_k$
\State \Return $(\pi_{k+1},P_k,H_k,\mathcal R_k)$
\end{algorithmic}
\end{algorithm}

\newcommand{\OuterHeader}[1]{\PhaseHeader{236,232,222}{#1}}
\newcommand{\LoopHeader}[1]{\PhaseHeader{228,236,228}{#1}}

\begin{algorithm}[t]
\caption{Multi-iteration RIPPLE with checkpointing and patience stopping}
\label{alg:ripple-full}
\small
\setlength{\baselineskip}{13pt}
\begin{algorithmic}[1]
\setlength{\itemsep}{1.2pt}
\Require Base policy $\pi_0$; training pool $\mathcal D_{\mathrm{train}}$;
         replay core $\mathcal D_{\mathrm{core}}$; replay mode $m$;
         patch library $\mathcal L$; aggregator $g$; rollout count $N$;
         diagnosis budget $B_{\mathrm{diag}}$; tolerances
         $(\varepsilon_R,\varepsilon_C)$; maximum iterations $K_{\max}$.
\Ensure Final policy $\pi_{\mathrm{final}}$ and ordered patch sequence $P$.

\OuterHeader{Initialize persistent state}
\State $P\gets()$; $A_{<0}\gets\varnothing$; $H_{-1}\gets\varnothing$;
       $u\gets0$ \Comment{consecutive no-update iterations}

\LoopHeader{Greedy outer-loop policy improvement}
\For{$k=0,\ldots,K_{\max}-1$}
    \State $(\pi_{k+1},P_k,H_k,\mathcal R_k)\gets
       \textsc{RIPPLE-Iter}\big(\pi_k,\mathcal D_{\mathrm{train}},
       \mathcal D_{\mathrm{core}},m,H_{k-1},\mathcal L,$
       \Statex \hspace*{6.2em}$g,N,B_{\mathrm{diag}},\varepsilon_R,\varepsilon_C,A_{<k}\big)$
       \Comment{Algorithm~\ref{alg:ripple-single-iter}}
    \State $P\gets P\mathbin\Vert P_k$;
       $A_{<k+1}\gets A_{<k}\cup\{\operatorname{id}(a):a\in P_k\}$
    \State \textsc{PersistCheckpoint}$(k+1,\pi_{k+1},P,\mathcal R_k)$
    \If{$P_k=\varnothing$}
        \State $u\gets u+1$
    \Else
        \State $u\gets0$
    \EndIf
    \If{$u\geq2$}
        \State \textbf{break}
    \EndIf
\EndFor
\State $\pi_{\mathrm{final}}\gets\pi_0\oplus P$
\State \Return $(\pi_{\mathrm{final}},P)$
\end{algorithmic}
\end{algorithm}

\subsubsection{Reward Function}
\label{app:reward}

\paragraph{Workflow modification.} The modification reward is
\[
R_{\mathrm{mod}}(\tau)
=
0.3S(\tau)+0.3C(\tau)+0.2S(\tau)C(\tau)+0.1E(\tau)-0.1K(\tau).
\]
The interaction term rewards workflows that are both correct and service-validated; edit efficiency and execution cost provide lighter shaping.

\paragraph{Component definitions.} $S(\tau)\in\{0,1\}$ equals one only when the agent produces a candidate workflow with at least $\max(\lfloor n_{\mathrm{cor}}/2\rfloor,3)$ actions, where $n_{\mathrm{cor}}$ is the action count of the corrupted input, and a post-hoc create-and-validate operation succeeds. The temporary workflow is deleted immediately after evaluation.

$C(\tau)\in[0,1]$ is a directed field-level comparison with the ground truth, keyed by action identifier and covering action type, ground-truth parameter keys, and all base, conditional, and error transitions. Missing actions mismatch all corresponding fields; extra actions receive a fixed three-field penalty; empty-workflow edge cases are handled explicitly.

$E(\tau)=1-\operatorname{UCR}(\tau)$, where the unnecessary-change ratio counts generated actions absent from both the corrupted and ground-truth workflows, edits to actions that were already correct, and deletions of ground-truth actions. A trajectory with no edits receives $E=1$; success and correctness separately penalize failure to implement the request.

Execution cost combines interaction and token components:
\[
\begin{aligned}
I(\tau)
&=
\alpha\min\!\left(\frac{\operatorname{turns}(\tau)}{T_{\max}},1\right)
+(1-\alpha)\min\!\left(\frac{\operatorname{tools}(\tau)}{U_{\max}},1\right),\\
K(\tau)
&=
\beta\min\!\left(\frac{\operatorname{tokens}(\tau)}{C_{\max}},1\right)
+(1-\beta)I(\tau),
\end{aligned}
\]
with $T_{\max}=20$, $U_{\max}=50$, $C_{\max}=100{,}000$, $\alpha=0.4$, and $\beta=0.7$. Rollouts are operationally capped at 15 turns; the larger normalization constants leave headroom below saturation.

\paragraph{Workflow generation.} Generation uses
\[
R_{\mathrm{gen}}(\tau)=C_{\mathrm{gen}}(\tau)-0.1K(\tau),
\]
where $C_{\mathrm{gen}}$ is identifier-invariant and combines square-root multiset-F1 scores over action types, edges, parameter keys, and normalized parameter values. Dynamically generated identifiers are replaced with placeholders before comparison. Service validation is reported separately but does not enter $R_{\mathrm{gen}}$; $E$ is fixed at one for bookkeeping.

\subsubsection{Rollout Configuration and Uncertainty}
\label{app:model-config}
\label{app:bootstrap-intervals}
\label{app:bootstrap}

\paragraph{Backbones and runtime.} Each experiment fixes both the hosted model endpoint and the tool-enabled runtime. Haiku~4.5 is the principal backbone; Gemma~3-12B-it and Ministral~3-14B are used only for cross-backbone transfer. All backbones share the same base policy $\pi_0$ and patch library $\mathcal L$.

\paragraph{User simulator.} When the agent asks a clarification question during a rollout, a deterministic user simulator answers on behalf of the requester. The simulator is a Python program that consults each task's ground-truth metadata for the missing-information slots and refusal conditions defined in the benchmark, and returns the corresponding answer verbatim; it does not paraphrase, does not volunteer information beyond what was asked, and makes no LLM calls. This lets the agent exercise the \texttt{CLARIFY} segment during training and diagnosis while keeping user responses reproducible across rollouts and backbones. The diagnosis module that maps trajectories to failure families is also a deterministic Python program.

\paragraph{Tools and environment.} The agent can read, write, edit, and search files; validate workflows through the same create-and-validate path used to define $S$; and resolve environment resources such as routing targets, functions, bots, prompts, schedules, reusable workflows, and integrations. Rollouts execute in a dedicated evaluation environment, and temporary validation resources are deleted after use.

\paragraph{Rollout execution and caching.} A rollout is capped at 15 agent turns, with per-turn and total timeouts. It terminates after a completed validation path or after two consecutive turns without tool calls, and one retry is permitted for a latency-only stall. We retain each endpoint's default sampling parameters. Rollouts are cached by $(\text{example\_id},\text{policy\_hash},\text{run\_idx})$, so any policy edit invalidates the corresponding cached trajectory.

\paragraph{Environment confounds.} Tool incompleteness, ambiguous resource descriptions, and systematic infrastructure failures are not modeled separately. They can therefore appear as F2 symptoms even when no prompt patch can correct the underlying environment issue.

\paragraph{Family-cluster bootstrap for Haiku deltas.} Haiku deltas are paired by Flow-HO datapoint. For $X\in\{S,C,R\}$, define
$\Delta X(d)=\bar X_{\mathrm{system}}(d)-\bar X_{\mathrm{ref}}(d)$,
where each bar averages the $N=3$ rollouts. The 39 paired differences are grouped into four ground-truth workflow families. Each of $10{,}000$ bootstrap replicates resamples four family labels with replacement, includes every paired datapoint from each sampled family, and computes the pooled datapoint mean. Repeated labels therefore repeat the entire cluster. The reported interval is the 2.5th--97.5th percentile range.

Resampling complete clusters preserves within-family dependence; because family sizes differ, the pooled statistic remains datapoint-weighted. The principal Haiku deltas and their family-cluster $95\%$ intervals are summarized in Table~\ref{tab:bootstrap-intervals}.
\begin{table}[H]
\centering
\setlength{\tabcolsep}{7pt}
\renewcommand{\arraystretch}{1.02}
\footnotesize
\begin{tabular}{@{}l r c@{}}
\toprule
\textbf{Metric} & \textbf{Paired $\Delta$ vs. $\pi_0$} & \textbf{$95\%$ interval} \\
\midrule
Service-validation success (pp) & $+23.1$ & $[+13,+34]$ \\
Workflow correctness (pp)       & $+0.8$  & $[-5,+8]$ \\
Composite reward                & $+0.119$ & $[+0.08,+0.20]$ \\
\bottomrule
\end{tabular}
\vspace{-5pt}
\caption{Family-cluster bootstrap $95\%$ intervals for the principal Haiku deltas relative to $\pi_0$ ($10{,}000$ replicates over four ground-truth workflow-family clusters).}
\label{tab:bootstrap-intervals}
\end{table}

As shown in Table~\ref{tab:bootstrap-intervals}, both the service-validation success gain ($+23.1$~pp) and the composite reward gain ($+0.119$) have intervals that exclude zero, whereas the workflow-correctness gain remains unresolved. With only four clusters, these intervals should be interpreted as descriptive family-cluster uncertainty for the current benchmark, not as a population-level guarantee or an independent confirmation result.

\subsection{Haiku Principal-Study Supplementary Results}
\label{app:supp-tables}
\label{app:haiku-supp}

This section provides the supplementary analyses for the Haiku RQ1--RQ4 study in \S\ref{sec:experiments}. Cross-backbone results for Gemma and Ministral appear in Appendix~\ref{app:cross-model-supp}.

\begin{table}[H]
\centering
\setlength{\tabcolsep}{5pt}
\renewcommand{\arraystretch}{1.06}
\footnotesize
\begin{tabular}{l ccccc}
\toprule
\textbf{Policy} & \textbf{$S$ (\%)} & \textbf{$C$ (\%)} & \textbf{$E$ (\%)} & \textbf{$K$ (\%)} & \textbf{$R$} \\
\midrule
Base $\pi_0$          & 54.70 & 74.87 & 50.44 & 8.53 & 0.5252 \\
Coverage-Adaptive     & 69.23 & 76.78 & 52.15 & 9.89 & 0.5992 \\
Train-Only            & 71.79 & 75.60 & 55.68 & 8.95 & 0.6132 \\
Multi-Iteration       & 64.10 & 74.33 & 51.78 & 9.35 & 0.5705 \\
\best{RIPPLE-U}       & \best{77.78} & 75.71 & \best{57.55} & \best{8.15} & \best{0.6443} \\
\best{RIPPLE-S}       & 76.07 & \best{77.88} & 56.47 & 9.48 & 0.6394 \\
Skill-Compact$^{\ddagger}$ & 70.94 & 82.18 & 54.57 & 8.76 & 0.6254 \\
\bottomrule
\end{tabular}
\vspace{-5pt}
\caption{Flow-HO modification reward components. $S,C,E,$ and $K$ are percentages; lower $K$ is better. $R$ uses the $[0,1]$ components in Appendix~\ref{app:reward}. $^{\ddagger}$Skill-Compact uses a separate acceptor-probe cache and is descriptive.}
\label{tab:reward-breakdown}
\end{table}

\subsubsection{Reward-component breakdown and family-level consistency}
\label{app:reward-breakdown}
\label{app:family-reward}

Table~\ref{tab:reward-breakdown} decomposes the Flow-HO modification reward across the harmonized Haiku systems. RIPPLE-U attains the highest validation success, edit efficiency, and composite reward, together with the lowest execution cost; RIPPLE-S attains the highest workflow correctness. Skill-Compact is included only as a descriptive checkpoint from the separate acceptor-probe cache.

\begin{figure}[H]
\centering
\includegraphics[width=0.82\linewidth]{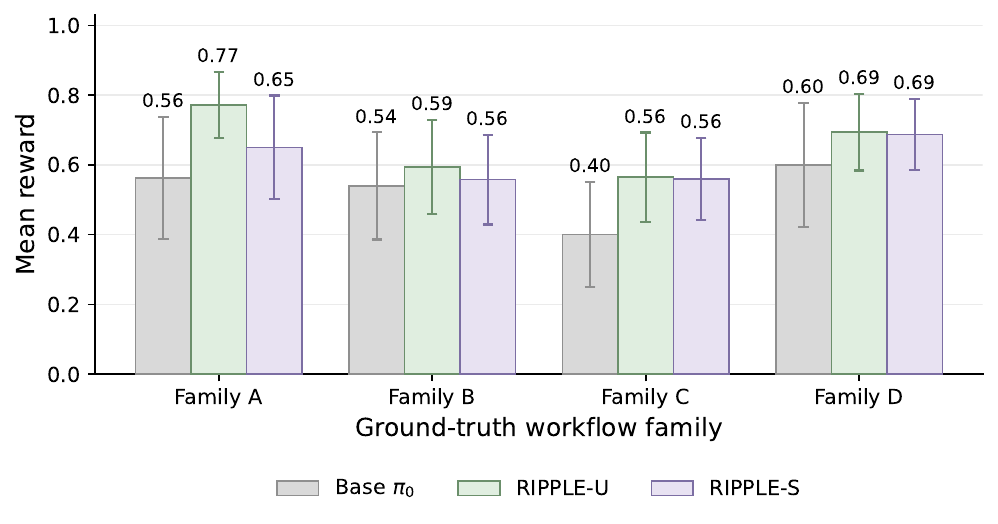}
\vspace{-15pt}
\caption{Mean reward by Flow-HO modification family in the Haiku study. Error bars show $1.96\times\mathrm{SEM}$ over task-level means; the comparison is descriptive because caches and rollout counts differ.}
\label{fig:per-family}
\end{figure}

Figure~\ref{fig:per-family} plots per-family mean reward with $1.96\times\mathrm{SEM}$ intervals, and Table~\ref{tab:family-reward} gives the underlying numeric means and rollout counts. Their source caches and rollout counts differ: the base uses $N=3$ from the acceptor-probe cache, whereas RIPPLE-U and RIPPLE-S use $N=10$ from the design-study cache. The comparison is therefore descriptive and supports family-level directionality, not paired inference.

\begin{table}[H]
\centering
\setlength{\tabcolsep}{6pt}
\renewcommand{\arraystretch}{1.15}
\footnotesize
\begin{tabular}{c c c c c}
\toprule
\textbf{Family} & $n$ & \textbf{Base} & \textbf{RIPPLE-U} & \textbf{RIPPLE-S} \\
\midrule
A & 9  & $0.563 \pm 0.175$ & $0.772 \pm 0.095$ & $0.650 \pm 0.149$ \\
B & 10 & $0.540 \pm 0.154$ & $0.594 \pm 0.135$ & $0.558 \pm 0.129$ \\
C & 10 & $0.401 \pm 0.151$ & $0.565 \pm 0.128$ & $0.560 \pm 0.117$ \\
D & 10 & $0.600 \pm 0.178$ & $0.694 \pm 0.110$ & $0.687 \pm 0.102$ \\
\bottomrule
\end{tabular}
\vspace{-5pt}
\caption{Per-family reward means with $1.96\times\mathrm{SEM}$ intervals. Base uses $N=3$ from the acceptor-probe cache; RIPPLE-U and RIPPLE-S use $N=10$ from the design-study cache, so the intervals are descriptive rather than harmonized and paired.}
\label{tab:family-reward}
\end{table}

\begin{figure}[H]
\centering
\includegraphics[width=0.82\linewidth]{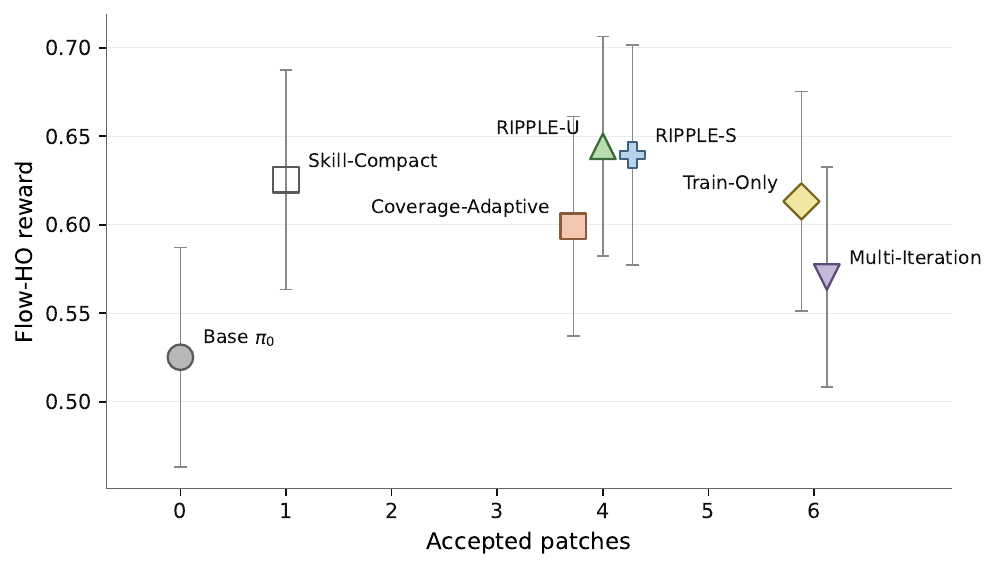}
\vspace{-10pt}
\caption{Accepted-patch count versus Flow-HO modification reward. Error bars show $\pm1$ SEM across 39 held-out tasks; points with equal counts are offset only for visibility. Skill-Compact is descriptive because it uses the acceptor-probe cache.}
\label{fig:pareto}
\end{figure}

\begin{figure}[H]
\centering
\includegraphics[width=0.98\textwidth]{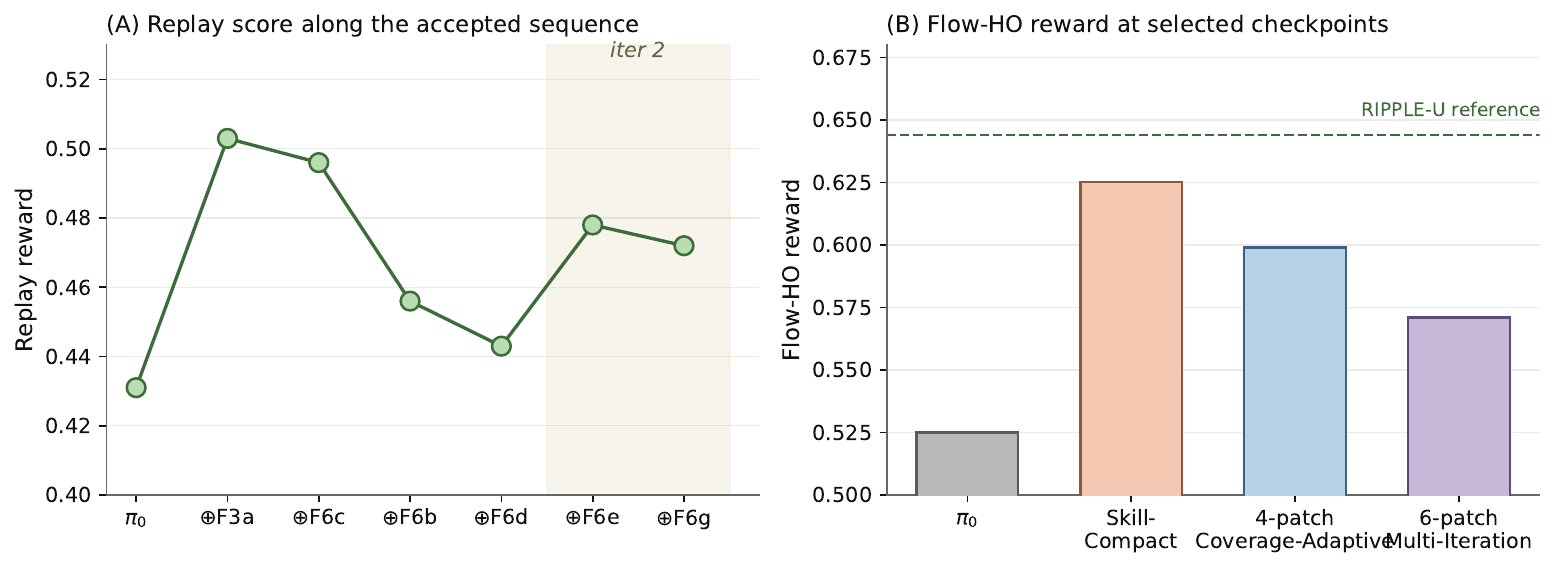}
\vspace{-10pt}
\caption{Coverage-Adaptive convergence. (A) Replay score after each accepted patch; shading marks iteration-2 acceptances. (B) Flow-HO scores at selected checkpoints. The dashed RIPPLE-U line follows a different four-patch sequence.}
\label{fig:convergence}
\end{figure}

\subsubsection{Convergence and checkpoint complementarity}
\label{app:convergence-diagnostics}
\label{app:convergence-tables}
\label{app:checkpoint-complementarity}

Figure~\ref{fig:pareto} places the harmonized Haiku systems on the patch-count--reward plane. Equal-size patch sets occupy different quality regimes, and the six-patch Multi-Iteration policy trails the principal variants. Figure~\ref{fig:convergence} traces Coverage-Adaptive through iterations~1 and~2; the dashed line shows the distinct four-patch RIPPLE-U sequence. Table~\ref{tab:convergence-replay} records the replay score after each acceptance in the Coverage-Adaptive and RIPPLE-U trajectories, including the two iteration-2 additions (F6e, F6g); Table~\ref{tab:convergence-flowho} reports the paired Flow-HO reward, correctness, and validation-success values at the same checkpoints. Together they show that iteration-2 acceptances raise the replay score but do not carry over to Flow-HO.

\begin{table}[H]
\centering
\setlength{\tabcolsep}{6pt}
\renewcommand{\arraystretch}{1.15}
\footnotesize
\begin{tabular}{l c c c}
\toprule
\textbf{Checkpoint} & \textbf{Success} & \textbf{Correctness} & \textbf{Reward} \\
\midrule
$\pi_0$                                              & 0.396 & 0.717 & 0.431 \\
$\pi_0\oplus\mathrm{F3a}$                            & 0.521 & 0.725 & 0.503 \\
$\cdots\oplus\mathrm{F6c}$                           & 0.500 & 0.736 & 0.496 \\
$\cdots\oplus\mathrm{F6b}$                           & 0.479 & 0.684 & 0.456 \\
$\cdots\oplus\mathrm{F6d}$ (Coverage-Adaptive final) & 0.354 & 0.776 & 0.443 \\
$\cdots\oplus\mathrm{F6e}$ (iter 2)                  & 0.533 & 0.710 & 0.478 \\
$\cdots\oplus\mathrm{F6g}$ (iter 2, Multi-Iter final)& 0.489 & 0.741 & 0.472 \\
\bottomrule
\end{tabular}
\caption{Replay scores along the Coverage-Adaptive accepted sequence (Figure~\ref{fig:convergence}A). Each row is the advancing baseline after the listed patch is accepted.}
\label{tab:convergence-replay}
\end{table}

\begin{table}[H]
\centering
\setlength{\tabcolsep}{6pt}
\renewcommand{\arraystretch}{1.15}
\footnotesize
\begin{tabular}{l c c c c}
\toprule
\textbf{Checkpoint} & \textbf{Success} & \textbf{Correctness} & \textbf{Reward} & $n$ \\
\midrule
$\pi_0$                                              & 0.547 & 0.749 & 0.525 & 39 \\
$\pi_0\oplus\mathrm{F3a}$ (Skill-Compact)           & 0.709 & 0.822 & 0.625 & 39 \\
4-patch (Coverage-Adaptive)                          & 0.692 & 0.768 & 0.599 & 39 \\
6-patch (Multi-Iteration)                            & 0.641 & 0.743 & 0.571 & 39 \\
\midrule
\textit{RIPPLE-U reference (Table~\ref{tab:main})}   & \textit{0.778} & \textit{0.757} & \textit{0.644} & \textit{39} \\
\bottomrule
\end{tabular}
\caption{Flow-HO modification scores at selected Coverage-Adaptive checkpoints (Figure~\ref{fig:convergence}B; $n=39$, $N=3$). RIPPLE-U is a reference with a different four-patch sequence.}
\label{tab:convergence-flowho}
\end{table}

\begin{figure}[H]
\centering
\includegraphics[width=0.76\linewidth]{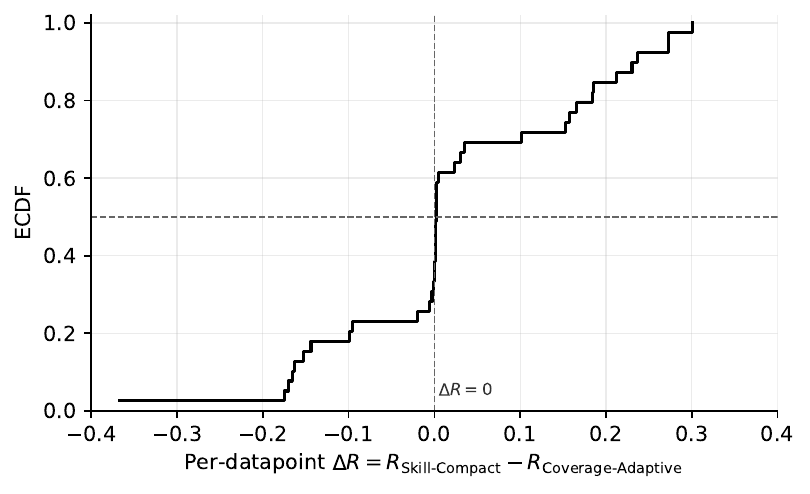}
\vspace{-10pt}
\caption{ECDF of task-level reward differences, Skill-Compact minus Coverage-Adaptive ($n=39$). Positive values favor the compact checkpoint; the sign change shows that neither checkpoint dominates all requests.}
\label{fig:ecdf}
\end{figure}

Figure~\ref{fig:ecdf} shows request-level complementarity between Skill-Compact and Coverage-Adaptive. Under strict sign counting, Skill-Compact is higher on $67\%$ of tasks. With the predefined $|\Delta R|\leq0.02$ tie band, the counts are 15 compact wins, 9 broader-checkpoint wins, and 15 ties. A post hoc oracle that chooses the better checkpoint per request raises mean reward from $0.625$ to $0.665$; this is an upper bound, not a deployable routing rule. A request-only selector using $q$ and $y_0$ recovers none of this headroom under leave-one-family-out evaluation. A diagnostic selector recovers $61\%$ but observes the ground-truth workflow and is therefore nondeployable.

\begin{table}[H]
\centering
\setlength{\tabcolsep}{4pt}
\renewcommand{\arraystretch}{1.06}
\footnotesize
\begin{tabular}{l|ccccc}
\toprule
\textbf{Patch} & \textbf{Cov-Adapt} & \textbf{RIPPLE-U} & \textbf{RIPPLE-S} & \textbf{Train-Only} & \textbf{Multi-Iter} \\
\midrule
F1a & $-\Delta$ & $-\Delta$ & $-\Delta$ & $-\Delta$ & $-\Delta$ \\
F2a & $\times$ & \textbf{A} & \textbf{A} & \textbf{A} & $-\Delta$ \\
F2b &           & $-\Delta$ &           &           &           \\
F3a & \textbf{A} & \textbf{A} & $\times$ & \textbf{A} & \textbf{A} \\
F3b &           &           &           &           & $-\Delta$ \\
F3c &           & $-\Delta$ &           &           &           \\
F5a & $\times$ & $\times$ & \textbf{A} & \textbf{A} & $-\Delta$ \\
F5b &           & \textbf{A} &           &           &           \\
F6a & $-\Delta$ &           & $-\Delta$ & $-\Delta$ & $-\Delta$ \\
F6b & \textbf{A} & \textbf{A} & $\times$ & \textbf{A} & \textbf{A} \\
F6c & \textbf{A} &           & \textbf{A} & \textbf{A} & \textbf{A} \\
F6d & \textbf{A} &           & \textbf{A} & \textbf{A} & \textbf{A} \\
F6e &           &           &           &           & \textbf{A} (iter 2) \\
F6f &           &           &           &           & $-\Delta$ \\
F6g &           &           &           &           & \textbf{A} (iter 2) \\
\midrule
\textbf{Accepted total} & \textbf{4} & \textbf{4} & \textbf{4} & \textbf{6} & \textbf{6} \\
\bottomrule
\end{tabular}
\vspace{-5pt}
\caption{Patch outcomes by configuration. \textbf{A}: accepted; $\times$: replay rejection; $-\Delta$: training pre-filter rejection; blank: not proposed. Train-Only bypasses replay promotion.}
\label{tab:patch-profile}
\end{table}

\subsubsection{Cross-configuration patch profile}
\label{app:patch-profile}

Table~\ref{tab:patch-profile} records each proposed patch and its terminal status across Haiku configurations. The three one-iteration gated variants each accept four patches, but the sets differ: F3a is accepted by Coverage-Adaptive and RIPPLE-U but rejected under static replay; F2a is rejected by Coverage-Adaptive but accepted by RIPPLE-U and RIPPLE-S; and F5a is accepted only by RIPPLE-S among the gated one-iteration variants. Train-Only accepts six patches because replay cannot remove positive-training-gain candidates. Multi-Iteration retains the four Coverage-Adaptive patches and adds F6e and F6g in iteration~2. Similar patch counts therefore do not imply similar policies or held-out performance.

\subsection{Cross-Backbone Transfer Supplementary Results}
\label{app:cross-model-supp}

This section provides the supplementary tables for the Gemma and Ministral transfer runs summarized in \S\ref{sec:principal}. Both use the same $N=3$ rollout protocol and patch library as the Haiku study. Ministral additionally bypasses the training-side positive-gain pre-filter (Appendix~\ref{app:ministral-protocol}).

\subsubsection{Backbone tier bracket}
\label{app:tier-bracket}

Table~\ref{tab:tier-bracket} reports the base-$\pi_0$ pilot on six frozen backbones, evaluated on the same 39 Flow-HO modification tasks with $N=3$. For descriptive model selection, we define peer ($S\geq40\%$), upper-headroom ($30\%\leq S<40\%$), lower-headroom ($20\%\leq S<30\%$), and collapse ($S<20\%$) bands. The study selects one backbone from each non-collapse band. Collapse-band backbones are excluded because the base policy provides too little reliable execution capability for the present adaptation study.

\begin{table}[H]
\centering
\setlength{\tabcolsep}{5pt}
\renewcommand{\arraystretch}{1.06}
\footnotesize
\begin{tabular}{l l c c c c}
\toprule
\textbf{Backbone} & \textbf{Tier band} & \textbf{$S$ (\%)} & \textbf{$C$ (\%)} & \textbf{$R$} & \textbf{Selected} \\
\midrule
Haiku 4.5           & peer             & 54.7 & 74.9 & 0.525 & \cmark~(principal) \\
Gemma 3-12B-it      & upper-headroom   & 37.6 & 80.7 & 0.460 & \cmark \\
Ministral 3-14B     & lower-headroom   & 24.8 & 59.7 & 0.299 & \cmark \\
Llama 3.3 70B       & collapse         & 18.8 & 81.0 & 0.392 & -- \\
Nova Micro          & collapse         & 18.8 & 61.8 & 0.268 & -- \\
Llama 3.1 8B        & collapse         & 15.4 & 45.3 & 0.219 & -- \\
\bottomrule
\end{tabular}
\vspace{-5pt}
\caption{Base-$\pi_0$ pilot on six frozen backbones (39 Flow-HO tasks, $N=3$). Descriptive success bands are peer ($S\geq40\%$), upper-headroom ($30\%\leq S<40\%$), lower-headroom ($20\%\leq S<30\%$), and collapse ($S<20\%$); one backbone is selected from each non-collapse band.}
\label{tab:tier-bracket}
\end{table}

\subsubsection{Cross-backbone reward-component breakdown}
\label{app:reward-breakdown-cross}

Table~\ref{tab:reward-breakdown-cross} decomposes Flow-HO modification reward for Gemma and Ministral. On Gemma, validation success and reward increase while correctness, edit efficiency, and execution cost remain nearly unchanged. On Ministral, validation success and reward also increase and edit efficiency improves slightly; execution cost rises, primarily through the token component, as the accepted patches lengthen the segmented prompt.

\begin{table}[H]
\centering
\setlength{\tabcolsep}{5pt}
\renewcommand{\arraystretch}{1.06}
\footnotesize
\begin{tabular}{l l ccccc}
\toprule
\textbf{Backbone} & \textbf{Policy} & \textbf{$S$ (\%)} & \textbf{$C$ (\%)} & \textbf{$E$ (\%)} & \textbf{$K$ (\%)} & \textbf{$R$} \\
\midrule
Gemma 12B     & Base $\pi_0$ & 37.61 & 80.68 & 48.31 & 8.01  & 0.4597 \\
Gemma 12B     & RIPPLE       & \best{44.44} & 80.68 & \best{48.53} & 8.19  & \best{0.4916} \\
\midrule
Ministral 14B & Base $\pi_0$ & 24.79 & \best{59.68} & 32.68 & \best{30.17} & 0.2992 \\
Ministral 14B & RIPPLE       & \best{31.62} & 59.20 & \best{33.94} & 34.63 & \best{0.3266} \\
\bottomrule
\end{tabular}
\vspace{-5pt}
\caption{Cross-backbone reward components on Flow-HO modification ($n=39$, $N=3$). $S,C,E,$ and $K$ are percentages; lower $K$ is better. $R$ follows Appendix~\ref{app:reward}, and \best{bold} marks the better mean within each backbone.}
\label{tab:reward-breakdown-cross}
\end{table}

\subsubsection{Cross-backbone family-level consistency}
\label{app:family-reward-cross}

Table~\ref{tab:family-reward-cross} reports means for the same four Flow-HO modification families used in the Haiku analysis. Validation success improves in three of four families on each backbone, and the regressing family differs: family C for Gemma and family B for Ministral. Family D shows the largest reward gain on both backbones, consistent with recovering headroom where the base policy is weakest. Because each transfer run uses one frozen protocol, these values support descriptive family-level directionality rather than paired inference.

\begin{table}[H]
\centering
\setlength{\tabcolsep}{5pt}
\renewcommand{\arraystretch}{1.06}
\footnotesize
\begin{tabular}{l c cc c cc c cc}
\toprule
& & \multicolumn{2}{c}{\textbf{Base $\pi_0$}} & & \multicolumn{2}{c}{\textbf{RIPPLE}} & & \multicolumn{2}{c}{\textbf{Paired $\Delta$}} \\
\cmidrule(lr){3-4}\cmidrule(lr){6-7}\cmidrule(lr){9-10}
\textbf{Backbone} & \textbf{Family} & $S$ & $R$ & & $S$ & $R$ & & $\Delta S$ & $\Delta R$ \\
\midrule
Gemma 12B & A ($n{=}9$)  & 0.556 & 0.593 && 0.667 & 0.642 && $+0.111$ & $+0.049$ \\
Gemma 12B & B ($n{=}10$) & 0.467 & 0.459 && 0.600 & 0.505 && $+0.133$ & $+0.047$ \\
Gemma 12B & C ($n{=}10$) & 0.467 & 0.514 && 0.367 & 0.469 && $-0.100$ & $-0.046$ \\
Gemma 12B & D ($n{=}10$) & 0.033 & 0.286 && 0.167 & 0.366 && $+0.133$ & $+0.079$ \\
\cmidrule(l){1-10}
Gemma 12B & \textit{all}~($n{=}39$) & 0.376 & 0.460 && 0.444 & 0.492 && $+0.068$ & $+0.032$ \\
\midrule
Ministral 14B & A ($n{=}9$)  & 0.333 & 0.418 && 0.444 & 0.496 && $+0.111$ & $+0.078$ \\
Ministral 14B & B ($n{=}10$) & 0.300 & 0.278 && 0.200 & 0.212 && $-0.100$ & $-0.067$ \\
Ministral 14B & C ($n{=}10$) & 0.133 & 0.227 && 0.200 & 0.224 && $+0.067$ & $-0.003$ \\
Ministral 14B & D ($n{=}10$) & 0.233 & 0.285 && 0.433 & 0.392 && $+0.200$ & $+0.106$ \\
\cmidrule(l){1-10}
Ministral 14B & \textit{all}~($n{=}39$) & 0.248 & 0.299 && 0.316 & 0.327 && $+0.068$ & $+0.027$ \\
\bottomrule
\end{tabular}
\vspace{-5pt}
\caption{Per-family Flow-HO modification means for Gemma 12B and Ministral 14B ($N=3$). Validation success improves in three of four families on each backbone; the regressing family differs across backbones. Family D has the largest reward gain for both.}
\label{tab:family-reward-cross}
\end{table}

\subsubsection{Accepted patches and per-patch replay evidence}
\label{app:cross-model-patches}

Table~\ref{tab:cross-model-patches} lists accepted patches in promotion order. The transfer policies are largely backbone-specific: Gemma accepts one \texttt{PLAN} patch and two \texttt{EDIT} patches, whereas Ministral accepts one \texttt{REQ\_UNDERSTANDING} patch and two \texttt{EDIT} patches. The shared concentration on \texttt{EDIT} reflects recurring schema and parameter failures; the remaining segment differs with the diagnosed backbone-specific failure profile.

\begin{table}[H]
\centering
\setlength{\tabcolsep}{4pt}
\renewcommand{\arraystretch}{1.08}
\footnotesize
\begin{tabularx}{\linewidth}{@{}l l l Y@{}}
\toprule
\textbf{Backbone} & \textbf{Configuration} & \textbf{Accepted patches} & \textbf{Segment sequence} \\
\midrule
Haiku 4.5 & RIPPLE-U          & F2a, F3a, F5b, F6b & TOOL, EDIT, EDIT, REQ \\
Haiku 4.5 & RIPPLE-S          & F2a, F5a, F6c, F6d & TOOL, EDIT, TOOL, EDIT \\
Haiku 4.5 & Coverage-Adaptive & F3a, F6c, F6b, F6d & EDIT, TOOL, REQ, EDIT \\
\midrule
Gemma 3-12B & RIPPLE          & F6a, F3a, F3b       & PLAN, EDIT, EDIT \\
Ministral 14B$^{\ast}$ & RIPPLE & F6g, F3d, F3e    & REQ, EDIT, EDIT \\
\bottomrule
\end{tabularx}
\vspace{-5pt}
\caption{Accepted patches in promotion order. REQ and TOOL abbreviate \texttt{REQ\_UNDERSTANDING} and \texttt{TOOL\_USE}. $^{\ast}$Ministral bypasses the training pre-filter; see Appendix~\ref{app:ministral-protocol}.}
\label{tab:cross-model-patches}
\end{table}

Table~\ref{tab:cross-model-trajectory} reports the training and replay deltas behind these decisions. On Gemma, the replay gate rejects F5a in iteration~2 despite a positive $\Delta_{\mathrm{train}}$, paralleling the replay-side discrimination in the Haiku F2a case study (\S\ref{sec:interaction-rq}). On Ministral, the three highest-ranked candidates in the diagnosis-matched pool are evaluated after bypassing the training pre-filter; all three satisfy the unchanged two-signal replay gate.

\begin{table}[H]
\centering
\setlength{\tabcolsep}{3.5pt}
\renewcommand{\arraystretch}{1.06}
\scriptsize
\resizebox{\linewidth}{!}{%
\begin{tabular}{l c l l r r r r l}
\toprule
\textbf{Backbone} & \textbf{Iter.} & \textbf{Patch} & \textbf{Segment} & $\Delta_{\mathrm{train}}$ & $\Delta_R^g$ & $\Delta_C^g$ & $\Delta_S^g$ & \textbf{Outcome} \\
\midrule
Gemma 12B & 1 & F6a & \texttt{PLAN}   & $+0.066$ & $+0.013$ & $+0.029$ & $+0.000$ & accepted \\
Gemma 12B & 1 & F5a & \texttt{EDIT}   & $-0.016$ & --       & --       & --       & train pre-filter \\
Gemma 12B & 1 & F3a & \texttt{EDIT}   & $-0.079$ & --       & --       & --       & train pre-filter \\
\cmidrule(l){1-9}
Gemma 12B & 2 & F3a & \texttt{EDIT}   & $+0.032$ & $+0.038$ & $+0.004$ & $+0.077$ & accepted \\
Gemma 12B & 2 & F3b & \texttt{EDIT}   & $+0.007$ & $+0.082$ & $+0.033$ & $+0.154$ & accepted \\
Gemma 12B & 2 & F5a & \texttt{EDIT}   & $+0.006$ & $-0.074$ & --       & $-0.154$ & replay reject \\
\midrule
Ministral 14B$^{\ast}$ & 1 & F6g & \texttt{REQ\_UNDERSTANDING} & $-0.016$ & $+0.033$ & $+0.057$ & $+0.026$ & accepted \\
Ministral 14B$^{\ast}$ & 1 & F3d & \texttt{EDIT}              & $-0.035$ & $+0.007$ & $-0.031$ & $+0.026$ & accepted \\
Ministral 14B$^{\ast}$ & 1 & F3e & \texttt{EDIT}              & $-0.067$ & $+0.014$ & $+0.047$ & $+0.000$ & accepted \\
\bottomrule
\end{tabular}%
}
\vspace{-5pt}
\caption{Per-patch training and replay deltas for Gemma and Ministral. Training gain is measured against $\pi_k$ on the diagnosis pool; replay deltas use the current accepted prefix. Dashes denote uncomputed or unrecorded quantities. Gemma F5a is rejected by replay despite positive training gain. $^{\ast}$Ministral bypasses the training pre-filter (Appendix~\ref{app:ministral-protocol}); the replay gate is unchanged.}
\label{tab:cross-model-trajectory}
\end{table}

\subsubsection{Ministral protocol: bypassing the training pre-filter}
\label{app:ministral-protocol}

RIPPLE's persistence decision is the replay-side AND rule, $\Delta_R^g\geq-0.05$ and $\Delta_C^g\geq-0.10$. The training condition $\Delta_{\mathrm{train}}>0$ is a compute-saving pre-filter that avoids replay evaluation for candidates with a negative training estimate. Every accepted Haiku and Gemma patch already satisfies this condition, so the pre-filter does not change their accepted sets.

For Ministral, all six candidates examined across two candidate pools have training deltas in $[-0.10,0]$. With the pre-filter enabled, no candidate reaches replay and the policy remains $\pi_0$. In the reported run, we bypass the pre-filter for the final diagnosis-matched top three and apply the unchanged replay gate. F6g, F3d, and F3e are accepted sequentially with the deltas in Table~\ref{tab:ministral-replay}. The three candidates from the earlier pool were not reevaluated under the bypassed protocol, so the trace does not establish how replay would have classified them.

\begin{table}[H]
\centering
\setlength{\tabcolsep}{4.7pt}
\renewcommand{\arraystretch}{1.18}
\scriptsize
\begin{tabular}{l l r r r r l}
\toprule
\textbf{Patch} & \textbf{Segment} & $\boldsymbol{\Delta}_{\textbf{train}}$ & $\boldsymbol{\Delta}_{R}^{g}$ & $\boldsymbol{\Delta}_{C}^{g}$ & $\boldsymbol{\Delta}_{S}^{g}$ & \textbf{Outcome} \\
\midrule
\multicolumn{7}{c}{\textit{Reported diagnosis-matched pool; pre-filter bypassed}} \\
\midrule
F6g & REQ  & $-0.016$ & $+0.033$ & $+0.057$ & $+0.026$ & accepted \\
F3d & EDIT & $-0.035$ & $+0.007$ & $-0.031$ & $+0.026$ & accepted \\
F3e & EDIT & $-0.067$ & $+0.014$ & $+0.047$ & $+0.000$ & accepted \\
\midrule
\multicolumn{7}{c}{\textit{Earlier pool; pre-filter enabled}} \\
\midrule
F5a & EDIT & $-0.037$ & \multicolumn{3}{c}{not replay-evaluated} & screened \\
F2a & TOOL & $-0.048$ & \multicolumn{3}{c}{not replay-evaluated} & screened \\
F3a & EDIT & $-0.097$ & \multicolumn{3}{c}{not replay-evaluated} & screened \\
\bottomrule
\end{tabular}
\vspace{-5pt}
\caption{Ministral candidate outcomes. The reported run bypasses the positive-training-gain pre-filter for the final three candidates and retains the advancing replay gate. Earlier candidates were screened under the default pre-filter and were not replay-classified under the bypassed protocol. REQ and TOOL abbreviate \texttt{REQ\_UNDERSTANDING} and \texttt{TOOL\_USE}.}
\label{tab:ministral-replay}
\end{table}

\section{Discussion}
\label{app:discussion}

\subsection{Why the replay gate uses two signals}
\label{app:gate-design}

RIPPLE promotes a candidate only when replay reward and workflow correctness both remain within their tolerances. Composite reward alone is insufficient because gains in validation success, edit efficiency, or execution cost can compensate for a correctness loss. Conversely, a structurally closer workflow can still fail service validation. Separate constraints keep both dimensions visible rather than permitting unrestricted compensation within one scalar.

The gate uses an AND rule because each signal defines an independent admissibility condition; an OR rule would allow an arbitrarily large regression in one metric whenever the other passed. The thresholds are mildly negative rather than strict-improvement tests because replay effects are estimated from few rollouts on a small pool. The gate is therefore a bounded empirical safeguard, not a guarantee of monotone improvement.

\paragraph{Marginal rather than cumulative control.} Each decision is relative to the latest accepted prefix, not to the iteration-start policy. Small admissible regressions can therefore accumulate. In the Coverage-Adaptive trace, F3a raises replay reward from $0.431$ to $0.503$, after which the next three accepted patches reduce it to $0.496$, $0.456$, and $0.443$. Every marginal change remains above $\varepsilon_R=-0.05$, and the final checkpoint remains above the iteration-start policy, but it falls below the one-patch checkpoint. A cumulative budget or an additional constraint against $\pi_k$ would provide stronger control when sequence-level monotonicity is required.

\subsection{What ``coverage-weighted'' aggregation means}
\label{app:coverage-terminology}

The implementation label \emph{coverage-weighted} denotes $g_{\mathrm{balanced}}$: datapoints are averaged within each ground-truth workflow family, and represented families are then weighted equally. It does not refer to coverage over failure families or accepted patch types.

The two notions of family are distinct. A \textbf{ground-truth workflow family} groups structurally related tasks and defines replay aggregation and cluster resampling. A \textbf{failure family} (F1--F6) is inferred from trajectory evidence and indexes candidate patches. Candidate quotas and coverage-first ordering operate on failure families; $g_{\mathrm{balanced}}$ operates on workflow families.

\subsection{When the correctness constraint becomes binding}
\label{app:correctness-binding}

Across 20 logged candidate decisions in the four principal modification runs, seven candidates are rejected: two by reward alone and five by both reward and correctness. No candidate fails only the correctness condition. For this candidate set and these thresholds, a reward-only replay gate would therefore reproduce the same decisions. This differs from Train-Only, which removes replay promotion entirely.

In the present trace, the correctness constraint is a latent safeguard rather than an additional source of rejections. Its structural motivation remains: composite reward can tolerate a correctness decline when other components improve. Demonstrating incremental decision value, however, requires candidate trajectories on which the reward and correctness conditions disagree.

\section{Segmented Prompt Policy: Base Text and Accepted Deltas}
\label{app:policy-showcase}

For auditability, this section shows the segment-level changes in the four-patch Coverage-Adaptive checkpoint, whose accepted sequence is F3a, F6c, F6b, and F6d. The original $\pi_0$ text is preserved within each segment, and every patch appends one bounded instruction to its designated location. RIPPLE-U and RIPPLE-S select different four-patch sequences and are not shown here.

Only the modified segments---\texttt{REQ\_UNDERSTANDING}, \texttt{TOOL\_USE}, and \texttt{EDIT}---are included; the remaining four segments are identical to $\pi_0$. Service-specific names are replaced with functional aliases for presentation without changing the policy logic.

\subsection{REQ\_UNDERSTANDING (modified by F6b)}

F6b is appended to the unchanged request-understanding segment.

\begin{promptbox}[\texttt{REQ\_UNDERSTANDING}: Base Policy and RIPPLE Delta]{promptteal}
\promptsection{Base policy $\pi_0$}
\begin{promptverb}
[REQ_UNDERSTANDING_START]
When the user provides a flow JSON and a modification request:
1. Read the existing flow JSON carefully to understand its current structure
2. Identify the type of modification requested (add block, reroute, modify
   configuration, or replace logic)
3. Determine which blocks in the flow are likely affected by the request
4. Note the flow's overall topology: entry point, main paths, error paths,
   and terminal blocks
[REQ_UNDERSTANDING_END]
\end{promptverb}

\promptsection{Accepted patch F6b (appended)}
\begin{promptverb}
TRIGGER: when the request names multiple capabilities or features.
Enumerate every capability the request asks for and ensure each one maps
to at least one concrete element in your output with its required settings
populated. Treat an unaddressed capability as incomplete work, not an
optional extra.
\end{promptverb}
\end{promptbox}

\textbf{Effect.} F6b turns multi-capability requests into an explicit coverage checklist, reducing partial outputs.

\subsection{TOOL\_USE (modified by F6c)}

F6c adds a resolve-before-fill requirement to the existing tool-use contract.

\begin{promptbox}[\texttt{TOOL\_USE}: Base Policy and RIPPLE Delta]{promptteal}
\promptsection{Base policy $\pi_0$}
\begin{promptverb}
[TOOL_USE_START]
Use the available tools to resolve resource identifiers when needed:
- get_routing_targets: retrieve identifiers for routing blocks
- get_functions: retrieve function identifiers for function-call blocks
- get_bots / get_bot_aliases: retrieve conversational-bot configuration
- get_prompts: retrieve prompt identifiers for message blocks
- get_schedules: retrieve schedule identifiers for schedule-check blocks
- get_workflows: retrieve workflow identifiers for workflow-transfer blocks
- validate_workflow: validate the final output through the external validator
[TOOL_USE_END]
\end{promptverb}

\promptsection{Accepted patch F6c (appended)}
\begin{promptverb}
TRIGGER: when an element references an external resource (routing target, function,
bot, prompt, schedule). Resolve the real identifier or value for that
resource before filling the field; do not leave required values blank or
substitute a placeholder. If a value cannot be resolved, surface that
rather than emitting an empty field.
\end{promptverb}
\end{promptbox}

\textbf{Effect.} F6c requires concrete resource resolution before populating a referenced field and makes unresolved values explicit.

\subsection{EDIT (modified by F3a and F6d)}

The \texttt{EDIT} segment receives F3a and F6d in acceptance order. Because its base text is unchanged, only the appended instructions are shown.

\begin{promptbox}[\texttt{EDIT}: Accepted RIPPLE Deltas]{promptteal}
\promptsection{F3a: schema-complete action insertion}
\begin{promptverb}
Before adding an Action of a type that has not appeared earlier in the flow,
look up the required Parameters and Errors entries in the workflow-language knowledge base.
Include all required fields. Do not use placeholders for required fields.
\end{promptverb}

\promptsection{F6d: complete element details}
\begin{promptverb}
TRIGGER: when an element has been added but its details may be incomplete.
For each element, complete its outgoing connections (default, conditional,
and error branches), its metadata, and every required sub-field for its
type. A structurally present but under-specified element is unfinished.
\end{promptverb}
\end{promptbox}

\textbf{Effect.} F3a requires a schema lookup before inserting an unseen action type; F6d requires the resulting action, branches, metadata, and required subfields to be complete before editing terminates.

\subsection{Summary of policy delta}

All four updates are additive and segment-scoped: no base instruction, segment boundary, or surrounding knowledge scaffold is overwritten. The checkpoint can therefore be reviewed, attributed, and rolled back patch by patch.

\section{Case Study: A Locally Helpful Patch with a Destructive Interaction}
\label{app:interaction-case}

This case follows F2a on the Coverage-Adaptive trajectory. The patch instructs the agent to use resource-lookup tools only for newly introduced references and to preserve existing identifiers that appear valid. The rule is locally plausible because it can avoid redundant lookups, but its effect depends on the instructions already active in the policy.

\paragraph{Shared-parent training estimate.} F2a is evaluated against the iteration-start policy $\pi_k=\pi_0$ and receives $\Delta_{\mathrm{train}}=+0.0465$, passing the positive-gain pre-filter. This estimate supports local utility under the shared parent; it does not test the patch after composition.

\paragraph{Advancing-prefix replay estimate.} By promotion time, the accepted prefix is $(\mathrm{F3a},\mathrm{F6c},\mathrm{F6b},\mathrm{F6d})$. Appending F2a yields the marginal replay effects in Table~\ref{tab:f2a-marginal}.

\begin{table}[H]
\centering
\footnotesize
\begin{tabular}{l r r r}
\toprule
\textbf{Signal} & \textbf{4-patch policy} & \textbf{After F2a} & \textbf{Marginal} \\
\midrule
Composite reward & 0.4426 & 0.2170 & $\mathbf{-0.2257}$ \\
Workflow correctness & 0.7761 & 0.6543 & $\mathbf{-0.1218}$ \\
Service-validation success & 0.354 & 0.000 & $\mathbf{-0.354}$ \\
\bottomrule
\end{tabular}
\caption{Marginal replay effect of appending F2a to the four-patch Coverage-Adaptive prefix. Both promotion conditions ($\Delta_R^g\geq-0.05$ and $\Delta_C^g\geq-0.10$) fail, and service-validation success falls to zero.}
\label{tab:f2a-marginal}
\end{table}

RIPPLE therefore rejects F2a before it enters the persistent checkpoint.

\paragraph{Interaction mechanism.} F6c requires every referenced resource to be resolved to a concrete value. F2a narrows that instruction by exempting identifiers already present in the corrupted input. Such identifiers can be syntactically plausible yet stale in the evaluation environment; preserving them suppresses the revalidation that F6c would otherwise trigger, leading to publication failures.

\paragraph{What the case establishes.} The logged trace shows that F2a's estimated effect depends on policy context: it is positive under shared-parent training evaluation and destructive after the accepted prefix is active. An evaluation anchored to $\pi_0$ would answer a different question and would not directly test this composition. The trace does not establish F2a's isolated replay effect or the population frequency of the interaction; it demonstrates why commit-time evaluation should use the policy state in which the patch would persist.

\section{Future Work}
\label{app:future-work}

\paragraph{Stronger confirmation and gate analysis.} The clearest extensions are a request-disjoint confirmation pool, more ground-truth families, and repeated runs across independent environments. Harmonized reruns of Skill-Compact and the transfer backbones would permit paired uncertainty estimates. Threshold sweeps, leave-one-patch-out attribution, and cumulative replay budgets could separate proposal quality, marginal safety, and sequence-level regression; anytime-valid rules such as PACE~\citep{pace} could support sequential reuse of a development pool.

\paragraph{Broader tasks and full-system baselines.} Flow-HO can be complemented by public execution-graded environments such as ALFWorld, $\tau$-Bench, and SWE-bench Verified. Full comparisons with GRASP~\citep{grasp}, SkillGen~\citep{skillgen}, and SkillOpt~\citep{skillopt} would vary both proposer and acceptor beyond RQ4's controlled acceptor swap. Comparisons with GEPA~\citep{gepa}, MIPROv2~\citep{mipro}, OPRO~\citep{opro}, and TextGrad~\citep{textgrad} would quantify the value of diagnosis, segmentation, and replay-constrained persistence.

\paragraph{Initialization and model migration.} The study starts from one human-authored $\pi_0$. Sweeps over minimal, generated, and expert initial policies could test dependence on initialization and convergence to similar fixed points. A deployment study could carry an adapted checkpoint across backbone upgrades and identify patches that transfer, become redundant, or require reversal.

\paragraph{Backbone-specific tuning.}
Our transfer runs reuse Haiku's proposer, patch library, replay composition, and gate tolerances without backbone-specific adaptation. The resulting Gemma and Ministral gains therefore reflect transfer under a shared configuration rather than tuned per-backbone performance, and the residual Haiku-to-transfer gap should not be interpreted as a method ceiling. Backbone-specific patch libraries, replay composition, and re-calibrated $(\varepsilon_R,\varepsilon_C)$ tolerances may recover additional headroom; we leave this tuning budget unmeasured.

\end{document}